\documentclass[10pt,twocolumn,letterpaper]{article}

\usepackage{cvpr}
\usepackage{times}
\usepackage{epsfig}
\usepackage{graphicx}
\usepackage{amsmath}
\usepackage{amssymb}
\usepackage{subfigure}
\usepackage{color}
\usepackage{multirow}
\usepackage{graphicx}
\usepackage{makecell}
\usepackage{authblk}
\usepackage{pifont}
\newcommand{\cmark}{\ding{51}}%
\newcommand{\xmark}{\ding{55}}%
\makeatletter
\renewcommand\AB@affilsepx{ \ \ \ \ \ \ \ \ \  \protect\Affilfont}
\makeatother

\usepackage[breaklinks=true,bookmarks=false]{hyperref}

\cvprfinalcopy 

\def\cvprPaperID{1065} 

\begin{document}
	
\title{Selective Kernel Networks}
\author[1,2]{Xiang Li\thanks{Xiang Li and Jian Yang are with PCA Lab, Key Lab of Intelligent Perception and Systems for High-Dimensional Information of Ministry of Education, and Jiangsu Key Lab of Image and Video Understanding for Social Security, School of Computer Science and Engineering, Nanjing University of Science and Technology, China. Xiang Li is also a visiting scholar at Momenta. Email: xiang.li.implus@njust.edu.cn}}
\author[3,2]{Wenhai Wang\thanks{Wenhai Wang is with National Key Lab for Novel Software Technology, Nanjing University. He was an research intern at Momenta.}}
\author[4]{Xiaolin Hu\thanks{Xiaolin Hu is with the Tsinghua National Laboratory for Information Science and Technology (TNList) Department of Computer Science and Technology, Tsinghua University, China.}}
\author[1]{Jian Yang\thanks{Corresponding author.}}
\affil[1]{PCALab, Nanjing University of Science and Technology}
\affil[2]{Momenta}
\affil[3]{Nanjing University}
\affil[4]{Tsinghua University}

\maketitle
\begin{abstract}
	In standard Convolutional Neural Networks (CNNs), the receptive fields of artificial neurons in each layer are designed to share the same size. It is well-known in the neuroscience community that  the receptive field size of visual cortical neurons are modulated by the stimulus, which has been rarely considered in constructing CNNs. We propose a dynamic selection mechanism in CNNs that allows each neuron to adaptively adjust its receptive field size based on multiple scales of input information. A building block called Selective Kernel (SK) unit is designed, in which multiple branches with different kernel sizes are fused using softmax attention that is guided by the information in these branches. Different attentions on these branches yield different sizes of the effective receptive fields of neurons in the fusion layer. Multiple SK units are stacked to a deep network termed Selective Kernel Networks (SKNets). On the ImageNet and CIFAR benchmarks, we empirically show that SKNet outperforms the existing state-of-the-art architectures with lower model complexity. Detailed analyses show that the neurons in SKNet can capture target objects with different scales, which verifies the capability of neurons for adaptively adjusting their receptive field sizes according to the input. The code and models are available at https://github.com/implus/SKNet. 
\end{abstract}
	
\section{Introduction}
	The local receptive fields (RFs) of neurons in the primary visual cortex (V1) of cats \cite{hubel1962receptive} have inspired the construction of Convolutional Neural Networks (CNNs) \cite{lecun1989backpropagation} in the last century, and it continues to inspire mordern CNN structure construction. For instance, it is well-known that in the visual cortex, the RF sizes of neurons in the same area (e.g., V1 region) are different, which enables the neurons to collect multi-scale spatial information in the same processing stage. This mechanism has been widely adopted in recent Convolutional Neural Networks (CNNs). A typical example is InceptionNets \cite{szegedy2015going,ioffe2015batch,szegedy2016rethinking,szegedy2017inception}, in which a simple concatenation is designed to aggregate multi-scale information from, e.g., 3$\times$3, 5$\times$5, 7$\times$7 convolutional kernels inside the ``inception'' building block.
	
	However, some other RF properties of cortical neurons have not been emphasized in designing CNNs, and one such property is the adaptive changing of RF size. Numerous experimental evidences have suggested that the RF sizes of neurons in the visual cortex are not fixed, but modulated by the stimulus. The Classical RFs (CRFs) of neurons in the V1 region was discovered by Hubel and Wiesel \cite{hubel1962receptive}, as determined by single oriented bars. Later, many studies (e.g., \cite{nelson1978orientation}) found that the stimuli outside the CRF will also affect the responses of neurons. The neurons are said to have non-classical RFs (nCRFs). In addition, the size of nCRF is related to the contrast of the stimulus: the smaller the contrast, the larger the effective nCRF size \cite{sceniak1999contrast}. Surprisingly, by stimulating nCRF for a period of time, the CRF of the neuron is also enlarged after removing these stimuli \cite{pettet1992dynamic}. All of these experiments suggest that the RF sizes of neurons are not fixed but modulated by stimulus \cite{spillmann2015beyond}. Unfortunately, this property does not receive  much attention in constructing deep learning models. {Those models with multi-scale information in the same layer such as InceptionNets have an inherent mechanism to adjust the RF size of neurons in the next convolutional layer according to the contents of the input, because the next convolutional layer linearly aggregates multi-scale information from different branches. But that linear aggregation approach may be insufficient to provide neurons powerful adaptation ability. }

	In the paper, we present a nonlinear approach to aggregate information from multiple kernels to realize the adaptive RF sizes of neurons. We introduce a ``Selective Kernel'' (SK) convolution, which consists of a triplet of operators: \emph{Split, Fuse} and \emph{Select}. The \emph{Split} operator generates multiple paths with various kernel sizes which correspond to different RF sizes of neurons. The \emph{Fuse} operator combines and aggregates the information from multiple paths to obtain a global and comprehensive representation for selection weights. The \emph{Select} operator aggregates the feature maps of differently sized kernels according to the selection weights.

	\begin{figure*}[t]
		\begin{center}
			\setlength{\fboxrule}{0pt}
			\fbox{\includegraphics[width=\textwidth]{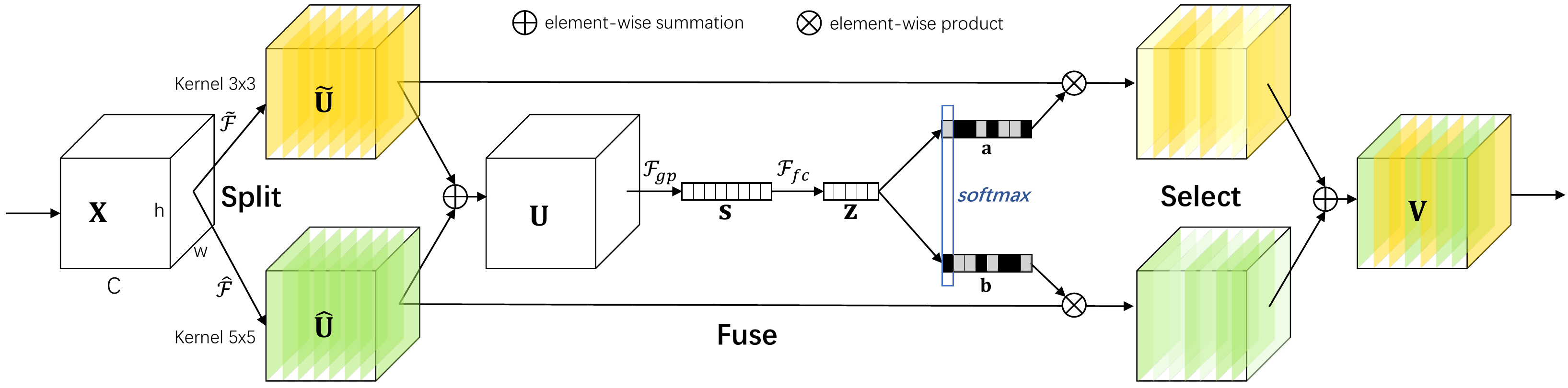}}
		\end{center}
		\vspace{-8pt}
		\caption{Selective Kernel Convolution.}	
		\label{fig_SK_cropped}
		\vspace{-9pt}
	\end{figure*}

	The SK convolutions can be computationally lightweight and impose only a slight increase in parameter and computational cost. We show that  on the ImageNet 2012 dataset  \cite{russakovsky2015imagenet} SKNets are superior to the previous state-of-the-art models with similar model complexity. Based on SKNet-50, we find the best settings for SK convolution and show the contribution of each component. To demonstrate their general applicability, we also provide compelling results on smaller datasets, CIFAR-10 and 100 \cite{krizhevsky2009learning}, and successfully embed SK into small models (e.g., ShuffleNetV2 \cite{ma2018shufflenet}).
	
	To verify the proposed model does have the ability to adjust neurons' RF sizes, we simulate the stimulus by enlarging the target object in natural images and shrinking the background to keep the image size unchanged. It is found that most neurons collect information more and more from the larger kernel path when the target object becomes larger and larger. These results suggest that the neurons in the proposed SKNet have adaptive RF sizes, which may underlie the model's superior performance in object recognition.
	
	\section{Related Work}
	\noindent \textbf{Multi-branch convolutional networks.} Highway networks \cite{srivastava2015highway} introduces the bypassing paths along with gating units. The two-branch architecture eases the difficulty to training networks with hundreds of layers. The idea is also used in ResNet \cite{he2016deep,he2016identity}, but the bypassing path is the pure identity mapping. Besides the identity mapping, the shake-shake networks \cite{gastaldi2017shake} and multi-residual networks \cite{abdi2016multi} extend the major transformation with more identical paths. The deep neural decision forests \cite{kontschieder2015deep} form the tree-structural multi-branch principle with learned splitting functions. FractalNets \cite{larsson2016fractalnet} and Multilevel ResNets \cite{zhang2017residual} are designed in such a way that the multiple paths can be expanded fractally and recursively. The InceptionNets \cite{szegedy2015going,ioffe2015batch,szegedy2016rethinking,szegedy2017inception} carefully configure each branch with customized kernel filters, in order to aggregate more informative and multifarious features. Please note that the proposed SKNets follow the idea of InceptionNets with various filters for multiple branches, but differ in at least two important aspects: 1) the schemes of SKNets are much simpler without heavy customized design and 2) an adaptive selection mechanism for these multiple branches is utilized to realize adaptive RF sizes of neurons.

	\noindent \textbf{Grouped/depthwise/dilated convolutions.} Grouped convolutions are becoming popular due to their low computational cost. Denote the group size by $G$, then both the number of parameters and the computational cost will be divided by $G$, compared to the ordinary convolution. They are first adopted in AlexNet \cite{krizhevsky2012imagenet} with a purpose of distributing the model over more GPU resources. Surprisingly, using grouped convolutions, ResNeXts \cite{xie2017aggregated} can also improve accuracy. This $G$ is called ``cardinality'', which characterize the model together with depth and width. 

	Many compact models such as IGCV1 \cite{zhang2017interleaved}, IGCV2 \cite{xie2018igcv} and IGCV3 \cite{sun2018igcv3} are developed, based on the interleaved grouped convolutions. A special case of grouped convolutions is depthwise convolution, where the number of groups is equal to the number of channels. Xception \cite{carreira1998xception} and MobileNetV1 \cite{howard2017mobilenets} introduce the depthwise separable convolution which decomposes ordinary convolutions into depthwise convolution and pointwise convolution. The effectiveness of depthwise convolutions is  validated in the subsequent works such as MobileNetV2 \cite{sandler2018mobilenetv2} and ShuffleNet \cite{zhang1707shufflenet,ma2018shufflenet}. Beyond grouped/depthwise convolutions, dilated convolutions \cite{yu2015multi,yu2017dilated} support exponential expansion of the RF without loss of coverage. For example, a 3$\times$3 convolution with dilation 2 can approximately cover the RF of a 5$\times$5 filter, whilst consuming less than half of the computation and memory. In SK convolutions, the kernels of larger sizes (e.g., $>$1) are designed to be integrated with the grouped/depthwise/dilated convolutions, in order to avoid the heavy overheads. 
	
	\noindent \textbf{Attention mechanisms.}  Recently, the benefits of attention mechanism have been shown across a range of tasks, from neural machine translation \cite{bahdanau2014neural} in natural language processing to image captioning \cite{you2016image} in image understanding. It biases the allocation of the most informative feature expressions \cite{itti2001computational,itti1998model,larochelle2010learning,mnih2014recurrent,olshausen1993neurobiological} and simultaneously suppresses the less useful ones. Attention has been widely used in recent applications such as person {re-ID} \cite{chen2018person}, image recovery \cite{zhang2018image}, text abstraction \cite{rush2015neural} and lip reading \cite{xu2018lcanet}. To boost the performance of image classification, Wang et al. \cite{wang2017residual} propose a trunk-and-mask attention between intermediate stages of a CNN. An hourglass module is introduced to achieve the global emphasis across both spatial and channel dimension. Furthermore, SENet \cite{hu2017squeeze} brings an effective, lightweight gating mechanism to self-recalibrate the feature map via channel-wise importances. Beyond channel, BAM \cite{park2018bam} and CBAM \cite{woo2018cbam} introduce spatial attention in a similar way. In contrast, our proposed SKNets are the first to explicitly focus on the adaptive RF size of neurons by introducing the attention mechanisms.
	
	\noindent \textbf{Dynamic convolutions.} Spatial Transform Networks \cite{jaderberg2015spatial} learns a parametric transformation to warp the feature map, which is considered difficult to be trained. Dynamic Filter \cite{jia2016dynamic} can only adaptively modify the parameters of filters, without the adjustment of kernel size. Active Convolution \cite{jeon2017active} augments the sampling locations in the convolution with offsets. These offsets are learned end-to-end but become static after training, while in SKNet the RF sizes of neurons can adaptively change during inference. Deformable Convolutional Networks \cite{dai2017deformable} further make the location offsets dynamic, {but it does not aggregate multi-scale information in the same way as SKNet does}.
	
	\begin{table*}
		\scriptsize
		\centering
		\renewcommand\arraystretch{1.1}
		\newcommand{\tabincell}[2]{\begin{tabular}{@{}#1@{}}#2\end{tabular}}
		
		\scalebox{1}{
			\begin{tabular}{c|c|c|c}
				\hline
				Output & ResNeXt-50 (32$\times$4d) & SENet-50 & SKNet-50 \\
				\hline
				112 $\times$ 112 &
				\multicolumn{3}{c}{7 $\times$ 7, 64, stride 2} \\ 
				\hline
				56 $\times$ 56 &
				\multicolumn{3}{c}{3 $\times$ 3 max pool, stride 2} \\
				\hline
				56 $\times$ 56
				&
				$
				\begin{bmatrix}
				\begin{array}{l}
				$1 $ \times $ 1$ , $ 128$  \\
				$3 $ \times $ 3$ , $ 128$, G = $ 32$  \\
				$1 $ \times $ 1$ , $ 256$  \\
				\end{array}
				\end{bmatrix} \times$ 3$
				$
				&
				$
				\begin{bmatrix}
				\begin{array}{l}
				$1 $ \times $ 1$ , $ 128$  \\
				$3 $ \times $ 3$ , $ 128$, G = $ 32$  \\
				$1 $ \times $ 1$ , $ 256$  \\
				fc, $ [16, 256]$
				\end{array}
				\end{bmatrix} \times$ 3$
				$ 
				&
				$
				\begin{bmatrix}
				\begin{array}{l}
				$1 $ \times $ 1$ , $ 128$  \\
				$SK[$M$ = 2, $G$ = 32, $r$ = 16], 128$    \\
				$1 $ \times $ 1$ , $ 256$  \\
				\end{array}
				\end{bmatrix} \times$ 3$
				$ 
				\\
				\hline
				28 $\times$ 28 &
				$
				\begin{bmatrix}
				\begin{array}{l}
				$1 $ \times $ 1$, $ 256$  \\
				$3 $ \times $ 3$, $ 256$, G = $ 32$ \\
				$1 $ \times $ 1$, $ 512$  \\
				\end{array}
				\end{bmatrix} \times$ 4$ 
				$
				&
				$
				\begin{bmatrix}
				\begin{array}{l}
				$1 $ \times $ 1$, $ 256$  \\
				$3 $ \times $ 3$, $ 256$, G = $ 32$ \\
				$1 $ \times $ 1$, $ 512$  \\
				fc, $ [32, 512]$\\
				\end{array}
				\end{bmatrix} \times$ 4$ 
				$ 
				&
				$
				\begin{bmatrix}
				\begin{array}{l}
				$1 $ \times $ 1$, $ 256$  \\
				$SK[$M$ = 2, $G$ = 32, $r$ = 16], 256$  \\
				$1 $ \times $ 1$, $ 512$  \\
				\end{array}
				\end{bmatrix} \times$ 4$
				$
				\\
				\hline
				14 $\times$ 14
				&
				$
				\begin{bmatrix}
				\begin{array}{l}
				$1 $\times$ 1$, $ 512$  \\
				$3 $\times$ 3$, $ 512$, G = $ 32$ \\
				$1 $\times$ 1$, $ 1024$ \\
				\end{array}
				\end{bmatrix} \times$ 6$
				$
				&
				$
				\begin{bmatrix}
				\begin{array}{l}
				$1 $\times$ 1$, $ 512$  \\
				$3 $\times$ 3$, $ 512$, G = $ 32$ \\
				$1 $\times$ 1$, $ 1024$ \\
				fc, $ [64, 1024]$\\
				\end{array}
				\end{bmatrix} \times$ 6$
				$
				&
				$
				\begin{bmatrix}
				\begin{array}{l}
				$1 $\times$ 1$, $ 512$  \\
				$SK[$M$ = 2, $G$ = 32, $r$ = 16], 512$  \\
				$1 $\times$ 1$, $ 1024$ \\
				\end{array}
				\end{bmatrix} \times$ 6$
				$
				\\
				\hline
				7 $\times$ 7 
				&
				$
				\begin{bmatrix}
				\begin{array}{l}
				$1 $\times$ 1$, $ 1024$  \\
				$3 $\times$ 3$, $ 1024$, G = $ 32$  \\
				$1 $\times$ 1$, $ 2048$ \\
				\end{array}
				\end{bmatrix} \times$ 3$
				$
				&
				$
				\begin{bmatrix}
				\begin{array}{l}
				$1 $\times$ 1$, $ 1024$  \\
				$3 $\times$ 3$, $ 1024$, G = $ 32$  \\
				$1 $\times$ 1$, $ 2048$ \\
				fc, $ [128, 2048]$\\
				\end{array}
				\end{bmatrix} \times$ 3$
				$
				&
				$
				\begin{bmatrix}
				\begin{array}{l}
				$1 $\times$ 1$, $ 1024$  \\
				$SK[$M$ = 2, $G$ = 32, $r$ = 16], 1024$   \\
				$1 $\times$ 1$, $ 2048$ \\
				\end{array}
				\end{bmatrix} \times 3
				$
				\\
				\hline
				1 $\times$ 1 & 
				\multicolumn{3}{c} {7 $\times$ 7 global average pool, 1000-d $fc$, {softmax}} \\
				\hline
				\#P  &25.0M &27.7M &27.5M \\
				\hline
				GFLOPs   &4.24 &4.25&4.47 \\
				\hline
		\end{tabular}}
		
		\vspace{+4pt}
		\caption{The three columns refer to ResNeXt-50 with a 32$\times$4d template, SENet-50 based on the ResNeXt-50 backbone and the corresponding SKNet-50, respectively. Inside the brackets are the general shape of a residual block, including filter sizes and feature dimensionalities. The number of stacked blocks on each stage is presented outside the brackets. ``$G=$ 32'' suggests the grouped convolution. The inner brackets following by $fc$ indicates the output dimension of the two fully connected layers in an SE module. 
			\#P denotes the number of parameter and the definition of FLOPs follow  \cite{zhang1707shufflenet}, i.e., the number of multiply-adds.}
		\label{tab_arch}
		\vspace{-2pt}
	\end{table*}
	
	\section{Methods}
	
	\subsection{Selective Kernel Convolution}
	To enable the neurons to adaptively adjust their RF sizes,  we propose an automatic selection operation, ``Selective Kernel'' (SK) convolution, among multiple kernels with different kernel sizes.
	Specifically, we implement the SK convolution via three operators -- \emph{Split}, \emph{Fuse} and \emph{Select}, as illustrated in Fig. \ref{fig_SK_cropped}, where a two-branch case is shown. Therefore in this example, there are only two kernels with different kernel sizes, but it is easy to extend to multiple branches case.
	
	\textbf{\emph{Split}}: For any given feature map $\mathbf{X} \in \mathbb{R}^{H' \times W' \times C'}$, by default we first conduct two transformations $\widetilde{\mathcal{F}}: \mathbf{X} \to \widetilde{\mathbf{U}} \in \mathbb{R}^{H \times W \times C}$ and $\widehat{\mathcal{F}}: \mathbf{X} \to \widehat{\mathbf{U}} \in \mathbb{R}^{H \times W \times C}$ with kernel sizes 3 and 5, respectively. Note that both $\widetilde{\mathcal{F}}$ and $\widehat{\mathcal{F}}$ are composed of efficient grouped/depthwise convolutions, Batch Normalization \cite{ioffe2015batch} and ReLU \cite{nair2010rectified} function in sequence. For further efficiency, the conventional convolution with a 5$\times$5 kernel is replaced with the dilated convolution with a 3$\times$3 kernel and dilation size 2. 
	
	\textbf{\emph{Fuse}}: As stated in Introduction, our goal is to enable neurons to adaptively adjust their RF sizes according to the stimulus content. The basic idea is to use gates to control the information flows from multiple branches carrying different scales of information into neurons in the next layer. To achieve this goal, the gates need to integrate information from all branches. We first fuse results from multiple (two in Fig. \ref{fig_SK_cropped}) branches via an element-wise summation: 
	\begin{equation}
	\mathbf{U} = \widetilde{\mathbf{U}} + \widehat{\mathbf{U}},
	\label{eq_u}
	\end{equation}
	then we embed the global information by simply using global average pooling to generate channel-wise statistics as $\mathbf{s} \in \mathbb{R}^C$. Specifically, the $c$-th element of $\mathbf{s}$ is calculated by shrinking $\mathbf{U}$ through spatial dimensions $H \times W$:
	\begin{equation}
	{s}_{c} = \mathcal{F}_{gp}(\mathbf{U}_{c}) = \frac{1}{H \times W}\sum_{i=1}^{H}\sum_{j=1}^{W} \mathbf{U}_{c}(i,j).
	\end{equation}
	
	Further, a compact feature $\mathbf{z} \in \mathbb{R}^{d \times 1}$ is created to enable the guidance for the precise and adaptive selections. This is achieved by a simple fully connected (fc) layer, with the reduction of dimensionality for better efficiency:
	\begin{equation}
	\mathbf{z} = \mathcal{F}_{fc}(\mathbf{s}) = \delta(\mathcal{B}(\mathbf{W} \mathbf{s})),
	\end{equation}
	where $\delta$ is the ReLU function \cite{nair2010rectified}, $\mathcal{B}$ denotes the Batch Normalization \cite{ioffe2015batch}, $\mathbf{W} \in \mathbb{R}^{d \times C}$. To study the impact of $d$ on the efficiency of the model, we use a reduction ratio $r$ to control its value:
	\begin{equation}\label{eqn:reduction_ratio}
	d=\max(C/r,L),
	\end{equation}
	where $L$ denotes the minimal value of $d$ ($L = $ 32 is a typical setting in our experiments).

	\textbf{\emph{Select}}: A soft attention across channels is used to adaptively select different spatial scales of information, which is guided by the compact feature descriptor $\mathbf{z}$. Specifically, a {softmax} operator is applied on the channel-wise digits:
	\begin{equation}
	{a}_c = \frac{e^{\mathbf{A}_c \mathbf{z}}}{e^{\mathbf{A}_c \mathbf{z}} + e^{\mathbf{B}_c \mathbf{z}}}, {b}_c = \frac{e^{\mathbf{B}_c \mathbf{z}}}{e^{\mathbf{A}_c \mathbf{z}} + e^{\mathbf{B}_c \mathbf{z}}}, 
	\label{eq_ab}
	\end{equation}
	where $\mathbf{A}, \mathbf{B} \in \mathbb{R}^{C \times d}$ and $\mathbf{a}, \mathbf{b}$ denote the soft attention vector for $\widetilde{\mathbf{U}}$ and $\widehat{\mathbf{U}}$, respectively. Note that $\mathbf{A}_c \in \mathbb{R}^{1 \times d}$ is the $c$-th row of $\mathbf{A}$ and ${a}_c$ is the $c$-th element of $\mathbf{a}$, likewise $\mathbf{B}_c$ and ${b}_c$. In the case of two branches, the matrix $\mathbf{B}$ is redundant because ${a}_c + {b}_c = 1$. The final feature map $\mathbf{V}$ is obtained through the attention weights on various kernels:
	\begin{equation}
	\mathbf{V}_c = {a}_c \cdot {\widetilde{\mathbf{U}}}_c + {b}_c \cdot \widehat{\mathbf{U}}_c, \ \ \ {a}_c + {b}_c = 1,
	\label{eq_v}
	\end{equation}
	where $\mathbf{V} = [\mathbf{V}_1, \mathbf{V}_2, ..., \mathbf{V}_C]$, $\mathbf{V}_c \in \mathbb{R}^{H \times W}$. Note that here we provide a formula for the two-branch case and one can easily deduce situations with more branches by extending Eqs.~\eqref{eq_u}~\eqref{eq_ab}~\eqref{eq_v}.

	\subsection{Network Architecture}\label{subsec:arch}
	Using the SK convolutions, the overall SKNet architecture is listed in Table \ref{tab_arch}. We start from ResNeXt \cite{xie2017aggregated} for two reasons: 1) it has low computational cost with extensive use of grouped convolution, and 2) it is one of the state-of-the-art network architectures with high performance on object recognition. Similar to the ResNeXt \cite{xie2017aggregated}, the proposed SKNet is mainly composed of a stack of repeated bottleneck blocks, which are termed ``SK units''. Each SK unit consists of a sequence of 1$\times$1 convolution, SK convolution and 1$\times$1 convolution. In general, all the large kernel convolutions in the original bottleneck blocks in ResNeXt are replaced by the proposed SK convolutions, enabling the network to choose appropriate RF sizes in an adaptive manner. As the SK convolutions are very efficient in our design, SKNet-50 only leads to 10\% increase in the number of parameters  and 5\% increase in computational cost, compared with ResNeXt-50.
	
	In SK units, there are three important hyper-parameters which determine the final settings of SK convolutions:  the number of paths $M$ that determines the number of choices of different kernels to be aggregated, the group number $G$ that controls the cardinality of each path, and the reduction ratio $r$ that controls the number of parameters in the {\it fuse} operator (see Eq. (\ref{eqn:reduction_ratio})). In Table \ref{tab_arch}, we denote one typical setting of SK convolutions SK[$M, G, r$] to be SK[2, 32, 16]. The choices and effects of these parameters are discussed in Sec. \ref{sec_ablation_imagenet}.
	
	{Table \ref{tab_arch} shows the structure of a 50-layer SKNet which has four stages with \{3,4,6,3\} SK units, respectively. By varying the number of SK units in each stage, one can obtain different architectures. In this study, we have experimented with other two architectures, SKNet-26, which has \{2,2,2,2\} SK units, and SKNet-101, which has \{3,4,23,3\} SK units, in their respective four stages.}
	
	Note that the proposed SK convolutions can be applied to other lightweight networks, e.g., MobileNet \cite{howard2017mobilenets,sandler2018mobilenetv2}, ShuffleNet \cite{zhang1707shufflenet,ma2018shufflenet}, in which 3$\times$3 depthwise convolutions are extensively used. By replacing these convolutions with the SK convolutions, we can also achieve very appealing results in the compact architectures (see Sec. \ref{exp_imagenet}).

	\section{Experiments}
	\subsection{ImageNet Classification}
	\label{exp_imagenet}
	The ImageNet 2012 dataset \cite{russakovsky2015imagenet} comprises 1.28 million training images and 50K validation images from 1,000 classes. We train networks on the training set and report the top-1 errors on the validation set. For data augmentation, we follow the standard practice and perform the random-size cropping to 224 $\times$224 and random horizontal flipping \cite{szegedy2015going}. The practical mean channel subtraction is adpoted to normalize the input images for both training and testing. Label-smoothing regularization \cite{szegedy2016rethinking} is used during training. For training large models, we use synchronous SGD with momentum 0.9, a mini-batch size of 256 and a weight decay of 1e-4. The initial learning rate is set to 0.1 and decreased by a factor of 10 every 30 epochs. All models are trained for 100 epochs from scratch on 8 GPUs, using the weight initialization strategy in \cite{he2015delving}. For training lightweight models, we set the weight decay to 4e-5 instead of 1e-4, and we also use slightly less aggressive scale augmentation for data preprocessing. Similar modifications can as well be referenced in \cite{howard2017mobilenets,zhang1707shufflenet} since such small networks usually suffer from underfitting rather than overfitting. To benchmark, we apply a centre crop on the validation set, where 224$\times$224 or 320$\times$320 pixels are cropped for evaluating the classification accuracy. The results reported on ImageNet are the averages of 3 runs by default. 
	
	\begin{table}
		\small
		\centering
		\renewcommand\arraystretch{1.1}
		\newcommand{\tabincell}[2]{\begin{tabular}{@{}#1@{}}#2\end{tabular}}
		\resizebox{0.5\textwidth}{!}{
			\begin{tabular}{l|c|c|c|c}
				\hline
				&\multicolumn{2}{c|}{top-1 err (\%)} & \multirow{2}{*}{\#P}   &\multirow{2}{*}{GFLOPs}   \\
				\cline{2-3}
				&224$\times$&320$\times$&&\\
				\hline
				ResNeXt-50 &22.23 &21.05&25.0M&4.24  \\
				AttentionNeXt-56 \cite{wang2017residual} &21.76&--&31.9M&6.32\\
				InceptionV3 \cite{szegedy2016rethinking} &-- & 21.20 & 27.1M & 5.73\\
				ResNeXt-50 + BAM \cite{park2018bam} &21.70&20.15&25.4M&4.31\\
				ResNeXt-50 + CBAM \cite{woo2018cbam} &21.40&20.38&27.7M&4.25\\
				SENet-50 \cite{hu2017squeeze}  &21.12&19.71 &{27.7}M&4.25  \\
				SKNet-50 (ours) &\textbf{20.79} &\textbf{19.32}&27.5M&{4.47}  \\
				\hline
				ResNeXt-101&21.11&19.86&44.3M& 7.99\\
				Attention-92 \cite{wang2017residual}&--&19.50&51.3M&10.43\\
				DPN-92 \cite{chen2017dual} &20.70&19.30& 37.7M &6.50\\
				DPN-98 \cite{chen2017dual} &20.20& 18.90& 61.6M &11.70\\
				InceptionV4 \cite{szegedy2017inception} &-- & 20.00 & 42.0M & 12.31\\
				Inception-ResNetV2 \cite{szegedy2017inception} &--&19.90&55.0M&13.22\\
				ResNeXt-101 + BAM \cite{park2018bam} &20.67&19.15&44.6M&8.05\\
				ResNeXt-101 + CBAM \cite{woo2018cbam}&20.60&19.42&49.2M&8.00\\
				SENet-101 \cite{hu2017squeeze} &20.58&18.61&{49.2}M&8.00\\
				SKNet-101 (ours)  &\textbf{20.19}&\textbf{18.40}&48.9M&{8.46}\\
				\hline
			\end{tabular}
		}
		\vspace{+1pt}
		
		\caption{Comparisons to the state-of-the-arts under roughly identical complexity. 224$\times$ denotes the single 224$\times$224 crop for evaluation, and likewise 320$\times$. Note that SENets/SKNets are all based on the corresponding ResNeXt backbones.}
		\label{tab_attention}
		\vspace{-2pt}
	\end{table}
	
	\vspace{6 pt}
	\noindent \textbf{Comparisons with state-of-the-art models.} We first compare SKNet-50 and SKNet-101 to the public competitive models with similar model complexity. The results show that SKNets consistently improve performance over the state-of-the-art attention-based CNNs under similar budgets. Remarkably, SKNet-50 outperforms ResNeXt-101 by above absolute 0.32\%, although ResNeXt-101 is 60\% larger in parameter and 80\% larger in computation. With comparable or less complexity than InceptionNets, SKNets achieve above absolute 1.5\% gain of performance, which demonstrates the superiority of adaptive aggregation for multiple kernels. We also note that using slightly less parameters, SKNets can obtain 0.3$\sim$0.4\% gains to SENet counterparts in both 224$\times$224 and 320$\times$320 evaluations. 
	
	\begin{table}
		\small
		\centering
		\renewcommand\arraystretch{1.1}
		\newcommand{\tabincell}[2]{\begin{tabular}{@{}#1@{}}#2\end{tabular}}
		\begin{tabular}{l|l|c|c}
			\hline
			&top-1 err. (\%) &  \#P & GFLOPs  \\
			\hline
			ResNeXt-50 (32$\times$4d) &22.23 &25.0M&4.24  \\
			\hline
			SKNet-50 (ours)   &\textbf{20.79} {\scriptsize(1.44)} &27.5M&{4.47}  \\
			\hline
			ResNeXt-50, {wider}&22.13 {\scriptsize(0.10)}&28.1M&4.74\\
			ResNeXt-{56}, deeper&22.04 {\scriptsize(0.19)}&27.3M&4.67\\
			ResNeXt-50 ({36}$\times$4d)&22.00 {\scriptsize(0.23)}&27.6M&4.70\\
			\hline
		\end{tabular}
		\vspace{+4pt}
		\caption{Comparisons on ImageNet validation set when the computational cost of model with more depth/width/cardinality is increased to match that of SKNet. The numbers in brackets denote the gains of performance.}
		\label{tab_compare_DWC}
		\vspace{-8pt}
	\end{table}

	\vspace{6 pt}
	\noindent \textbf{Selective Kernel \emph{vs.} Depth/Width/Cardinality.} Compared with ResNeXt (using the setting of 32$\times$4d), SKNets inevitably introduce a slightly increase in parameter and computation due to the additional paths of differnet kernels and the selection process. For fair comparison, we increase the complexity of ResNeXt by changing its depth, width and cardinality, to match the complexity of SKNets. Table \ref{tab_compare_DWC} shows that increased complexity does lead to better prediction accuracy.
	
	However, the improvement is marginal when going deeper (0.19\% from ResNeXt-50 to ResNeXt-53) or wider (0.1\% from ResNeXt-50 to ResNeXt-50 wider), or with slightly more cardinality  (0.23\% from ResNeXt-50 (32$\times$4d) to ResNeXt-50 (36$\times$4d)). 
	
	In contrast, SKNet-50 obtains 1.44\% absolute improvement over the baseline ResNeXt-50, which indicates that SK convolution is very efficient.
	
\begin{figure}[t]
		\begin{center}
			\setlength{\fboxrule}{0pt}
			\fbox{\includegraphics[width=0.44\textwidth]{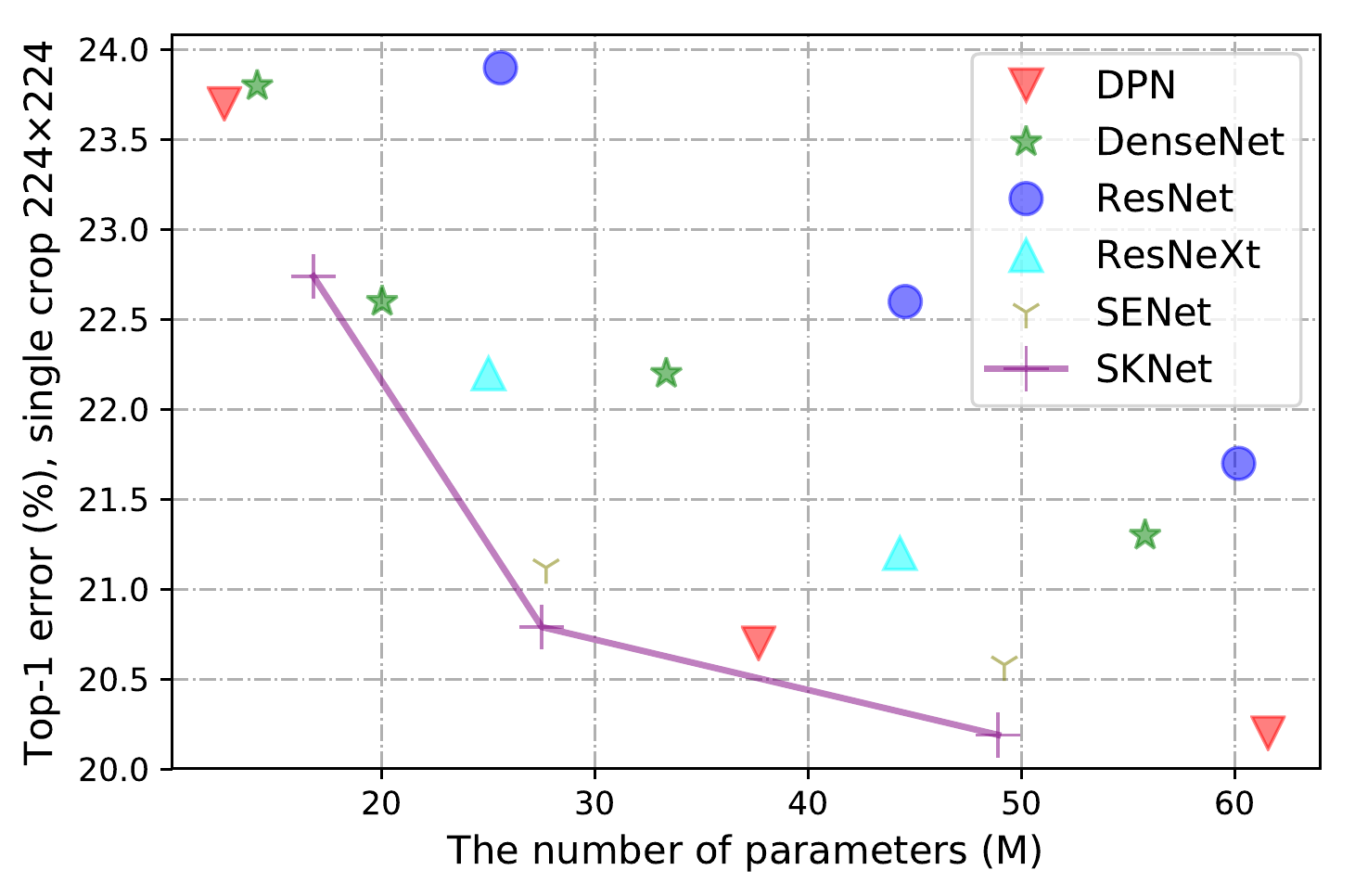}}
		\end{center}	
		\vspace{-14pt}
		\caption{Relationship between the performance of SKNet and the number of parameters in it, compared with the state-of-the-arts.}
		\label{fig_parameter}
		\vspace{-2pt}
	\end{figure}

	\vspace{6 pt}
	\noindent \textbf{Performance with respect to the number of parameters.} We plot the top-1 error rate of the proposed SKNet with respect to the number of parameters in it (Fig. \ref{fig_parameter}). Three architectures, SK-26, SKNet-50 and SKNet-101 (see Section \ref{subsec:arch} for details), are shown in the figure. For comparison, we plot the results of some state-of-the-art models including ResNets \cite{he2016deep}, ResNeXts \cite{xie2017aggregated}, DenseNets \cite{huang2017densely}, DPNs \cite{chen2017dual} and SENets \cite{hu2017squeeze} in the figure. Each model has multiple variants. The details of the compared architectures are provided in the Supplementary Materials. {All Top-1 errors are reported in the references.} It is seen that SKNets utilizes parameters more efficiently than these models. For instance, achieving $\sim$20.2 top-1 error, SKNet-101 needs 22\% fewer parameters than DPN-98. 
	
	\begin{table}
		\small
		\centering
		\renewcommand\arraystretch{1.1}
		\newcommand{\tabincell}[2]{\begin{tabular}{@{}#1@{}}#2\end{tabular}}
		\begin{tabular}{l|c|c|c}
			\hline
			ShuffleNetV2  & top-1 err.(\%) & MFLOPs&\#P \\
			\hline
			0.5$\times$ \cite{ma2018shufflenet} & 39.70 & 41 & 1.4M\\
			0.5$\times$ (our impl.) & 38.41 & 40.39 & 1.40M\\
			0.5$\times$ + SE \cite{hu2017squeeze} & 36.34 & 40.85 & 1.56M\\
			0.5$\times$ + SK & \textbf{35.35}&42.58&1.48M\\
			\hline
			1.0$\times$ \cite{ma2018shufflenet} & 30.60 & 146 & 2.3M\\
			1.0$\times$ (our impl.) & 30.57 & 140.35&2.45M \\
			1.0$\times$ + SE \cite{hu2017squeeze} &29.47 &141.73 &2.66M\\
			1.0$\times$ + SK &\textbf{28.36}&145.66&2.63M \\
			\hline
		\end{tabular}
		\vspace{+4pt}
		\caption{Single 224$\times$224 crop top-1 error rates (\%) by variants of lightweight models on ImageNet validation set.}
		\label{table_small_model}
		\vspace{-6pt}
	\end{table}
	
	\vspace{6 pt}
	\noindent \textbf{Lightweight models.} Finally, we choose the representative compact architecture -- ShuffleNetV2 \cite{ma2018shufflenet}, which is one of the strongest light models, to evaluate the generalization ability of SK convolutions. By exploring different scales of models in Table \ref{table_small_model}, we can observe that SK convolutions not only boost the accuracy of baselines significantly but also perform better than SE \cite{hu2017squeeze} (achieving around absolute 1\% gain). This indicates the great potential of the SK convolutions in applications on low-end devices.

	\subsection{CIFAR Classification}
	To evaluate the performance of SKNets on smaller datasets, we conduct more experiments on CIFAR-10 and 100 \cite{krizhevsky2009learning}. The two CIFAR datasets \cite{krizhevsky2009learning} consist of colored natural scence images, with 32$\times$32 pixel each. The train and test sets contain 50k images and 10k images respectively. CIFAR-10 has 10 classes and CIFAR-100 has 100. We take the architectures as in \cite{xie2017aggregated} for reference: our networks have a single 3$\times$3 convolutional layer, followed by 3 stages each having 3 residual blocks with SK convolution. We also apply SE blocks on the same backbone (ResNeXt-29, 16$\times$32d) for better comparisons. More architectural and training details are provided in the supplemantary materials. Notably, SKNet-29 achieves better or comparable performance than ResNeXt-29, 16$\times$64d with 60\% fewer parameters and it consistently outperforms SENet-29 on both CIFAR-10 and 100 with 22\% fewer parameters.
	\begin{table}
		\small
		\centering
		\renewcommand\arraystretch{1.1}
		\newcommand{\tabincell}[2]{\begin{tabular}{@{}#1@{}}#2\end{tabular}}
		\begin{tabular}{l|c|c|c}
			\hline

			Models& \#P & CIFAR-10 & CIFAR-100 \\
			\hline
			ResNeXt-29, 16$\times$32d &  25.2M & 3.87 & 18.56 \\
			ResNeXt-29, 8$\times$64d  & 34.4M & 3.65 & 17.77  \\
			ResNeXt-29, 16$\times$64d  & 68.1M & 3.58 & \textbf{17.31} \\
			\hline
			SENet-29 \cite{hu2017squeeze} & 35.0M & 3.68 &  17.78 \\
			SKNet-29 (ours) & 27.7M & \textbf{3.47} & 17.33\\
			\hline
		\end{tabular}
		\vspace{+4pt}
		\caption{Top-1 errors (\%, average of 10 runs) on CIFAR. SENet-29 and SKNet-29 are all based on ResNeXt-29, 16$\times$32d. }
		\label{tab_cifar}
		\vspace{-2pt}
	\end{table}

	\subsection{Ablation Studies}
	\label{sec_ablation_imagenet}
	In this section, we report ablation studies on the ImageNet dataset to investigate the effectiveness of SKNet.

	\begin{table}
		\small
		\centering
		\renewcommand\arraystretch{1.1}
		\newcommand{\tabincell}[2]{\begin{tabular}{@{}#1@{}}#2\end{tabular}}
		\begin{tabular}{c|c|c|l|c|c|c}
			\hline
			\multicolumn{3}{c|}{Settings} &\multirow{2}{*}{\makecell[c]{top-1\\err. (\%)}}& \multirow{2}{*}{\#P} &\multirow{2}{*}{GFLOPs}&\multirow{2}{*}{\makecell[c]{Resulted\\Kernel}}\\	
			\cline{1-3}
			Kernel & $D$ & $G$ &&&&\\
			\hline
			3$\times$3&3&32&20.97&27.5M&4.47&7$\times $7\\
			3$\times$3&2&32&\textbf{20.79}&27.5M&4.47&5$\times$5\\
			3$\times$3&1&32&20.91&27.5M&4.47&3$\times$3\\
			5$\times$5&1&64&\textbf{20.80}&28.1M&4.56&5$\times$5\\
			7$\times$7&1&128&21.18&28.1M&4.55&7$\times$7\\

			\hline
		\end{tabular}
		\vspace{+2pt}
		\caption{Results of SKNet-50 with different settings in the second branch, while the setting of the first kernel is fixed. ``Resulted kernel'' in the last column means the approximate kernel size with dilated convolution.}
		\label{table_dilation}
		\vspace{-8pt}
	\end{table}
	\begin{table}
		\small
		\centering
		\renewcommand\arraystretch{1.1}
		\newcommand{\tabincell}[2]{\begin{tabular}{@{}#1@{}}#2\end{tabular}}
		\begin{tabular}{c|c|c||c||c|c|c}
			\hline
			K3&K5&K7 & SK & \makecell[c]{top-1\\err. (\%)} & \#P & GFLOPs \\	
			\hline
			${\checkmark}$&&&&22.23 &25.0M&4.24\\
			&${\checkmark}$&&&25.14&25.0M&4.24\\
			&&${\checkmark}$&&25.51&25.0M&4.24\\
			\hline
			${\checkmark}$ &${\checkmark}$& &&21.76&26.5M&{4.46}\\
			${\checkmark}$ &${\checkmark}$& &${\checkmark}$&{20.79} &27.5M&{4.47}\\
			${\checkmark}$& &${\checkmark}$ &&21.82&26.5M&{4.46}\\
			${\checkmark}$& &${\checkmark}$ &${\checkmark}$&20.97&27.5M&{4.47}\\
			&${\checkmark}$&${\checkmark}$ &&23.64&26.5M&{4.46}\\
			&${\checkmark}$&${\checkmark}$ &${\checkmark}$&23.09&27.5M&{4.47}\\
			\hline
			${\checkmark}$ &${\checkmark}$ &${\checkmark}$& &21.47&28.0M&4.69\\
			${\checkmark}$ &${\checkmark}$ &${\checkmark}$&${\checkmark}$&20.76&29.3M&4.70\\
			\hline
		\end{tabular}
		\vspace{+4pt}
		\caption{Results of {SKNet-50} with different combinations of multiple kernels. Single 224$\times$224 crop is utilized for evaluation.}
		\label{table_C135}
		\vspace{-8pt}
	\end{table}

	\vspace{6 pt}
	\noindent \textbf{The dilation $D$ and group number $G$.} The dilation $D$ and group number $G$ are two crucial elements to control the RF size. To study their effects, we start from the two-branch case and fix the setting 3$\times$3 filter with dilation $D$ = 1 and group $G$ = 32 in the first kernel branch of SKNet-50. 

	Under the constraint of similar overall complexity, there are two ways to enlarge the RF of the second kernel branch: 1) increase the dilation $D$ whilst fixing the group number $G$, and 2) simultaneously increase the filter size and the group number $G$.
	
	Table \ref{table_dilation} shows that the optimal settings for the other branch are those with kernel size 5$\times$5 (the last column), which is larger than the first fixed kernel with size 3$\times$3. It is proved beneficial to use different kernel sizes, and we attribute the reason to the aggregation of multi-scale information.
	
	{There are two optimal configurations: kernel size 5$\times$5 with $D$ = 1 and kernel size 3$\times$3 with $D$ = 2}, where the latter has slightly lower model complexity. In general, we empirically find that the series of 3$\times$3 kernels with various dilations is moderately superior to the corresponding counterparts with the same RF (large kernels without dilations) in both performance and complexity.

	\vspace{6 pt}
	\noindent \textbf{Combination of different kernels.} Next we investigate the effect of combination of different kernels. Some kernels may have size larger than 3$\times$3, and there may be more than two kernels.  
	To limit the search space, we only use three different kernels, called ``K3'' (standard 3$\times$3 convolutional kernel), ``K5'' (3$\times$3 convolution with dilation 2 to approximate 5$\times$5 kernel size), and ``K7'' (3$\times$3 with dilation 3 to approximate 7$\times$7 kernel size). Note that  we only consider the dilated versions of large kernels (5$\times$5 and 7$\times$7) as Table \ref{table_dilation} has suggested. $G$ is fixed to 32. If ``SK'' in Table \ref{table_C135} is ticked, it means that we use the SK attention across the corresponding kernels {ticked in the same row} ({the output of each SK unit is} $\textbf{V}$ in Fig. \ref{fig_SK_cropped}), otherwise we simply sum up the results with these kernels ({then the output of each SK unit is} $\textbf{U}$ in Fig. \ref{fig_SK_cropped}) as a naive baseline model.

	The results in Table \ref{table_C135} indicate that excellent performance of SKNets can be attributed to the use of multiple kernels and the adaptive selection mechanism among them.
	From Table \ref{table_C135}, we have the following observations: (1)
	When the number of paths $M$ increases, in general the recognition error decreases. The top-1 errors in the first block of the table ($M$ = 1) are generally higher than those in the second block ($M$ = 2), and the errors in the second block are generally higher than the third block ($M$ = 3). 
	(2) No matter $M$ = 2 or 3, SK attention-based aggregation of multiple paths always achieves lower top-1 error than the simple aggregation method (naive baseline model). 
	(3) Using SK attention, the performance gain of the model from $M$ = 2 to $M$ = 3 is marginal (the top-1 error decreases from 20.79\% to 20.76\%). For better trade-off between performance and efficiency, $M$ = 2 is preferred.

	\begin{figure}[t]
		\begin{center}
			\setlength{\fboxrule}{0pt}
			\subfigure[]{
				\label{fig:subfig:a}\includegraphics[width=0.48\textwidth]{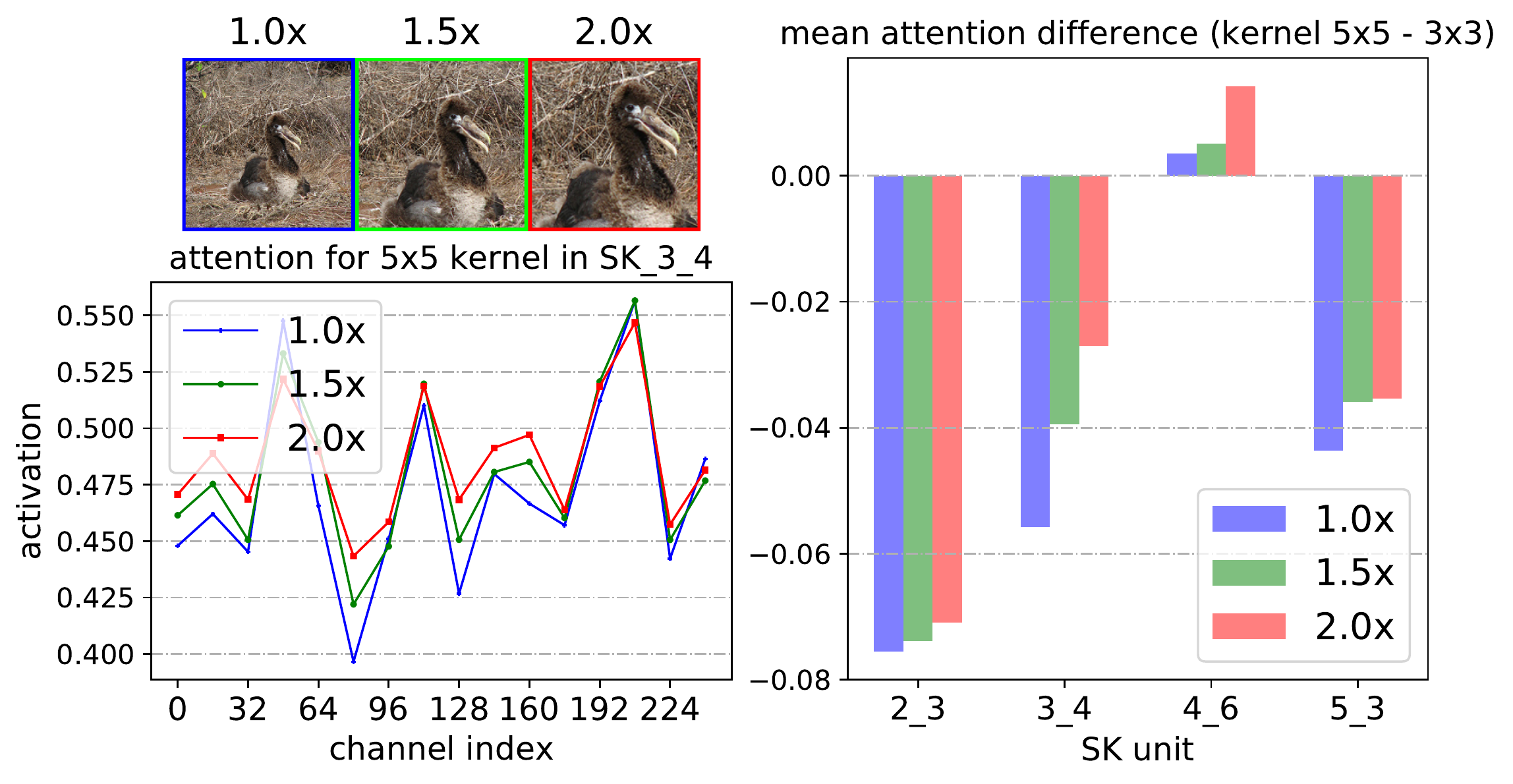}}
			\vspace{-4pt}
			\subfigure[]{
				\label{fig:subfig:b}\includegraphics[width=0.48\textwidth]{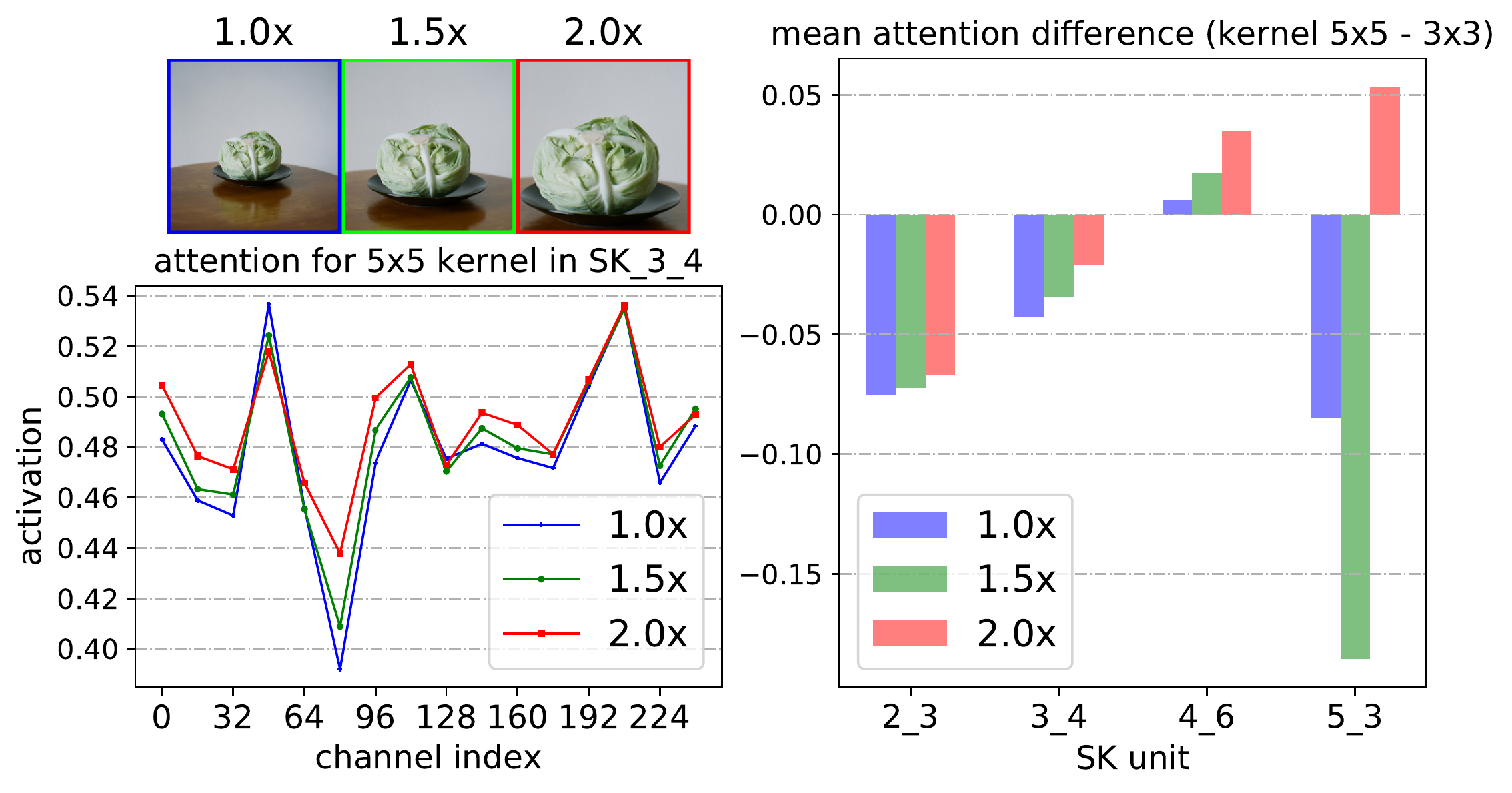}}
			\vspace{-4pt}
			\subfigure[]{
				\label{fig:subfig:c}\includegraphics[width=0.48\textwidth]{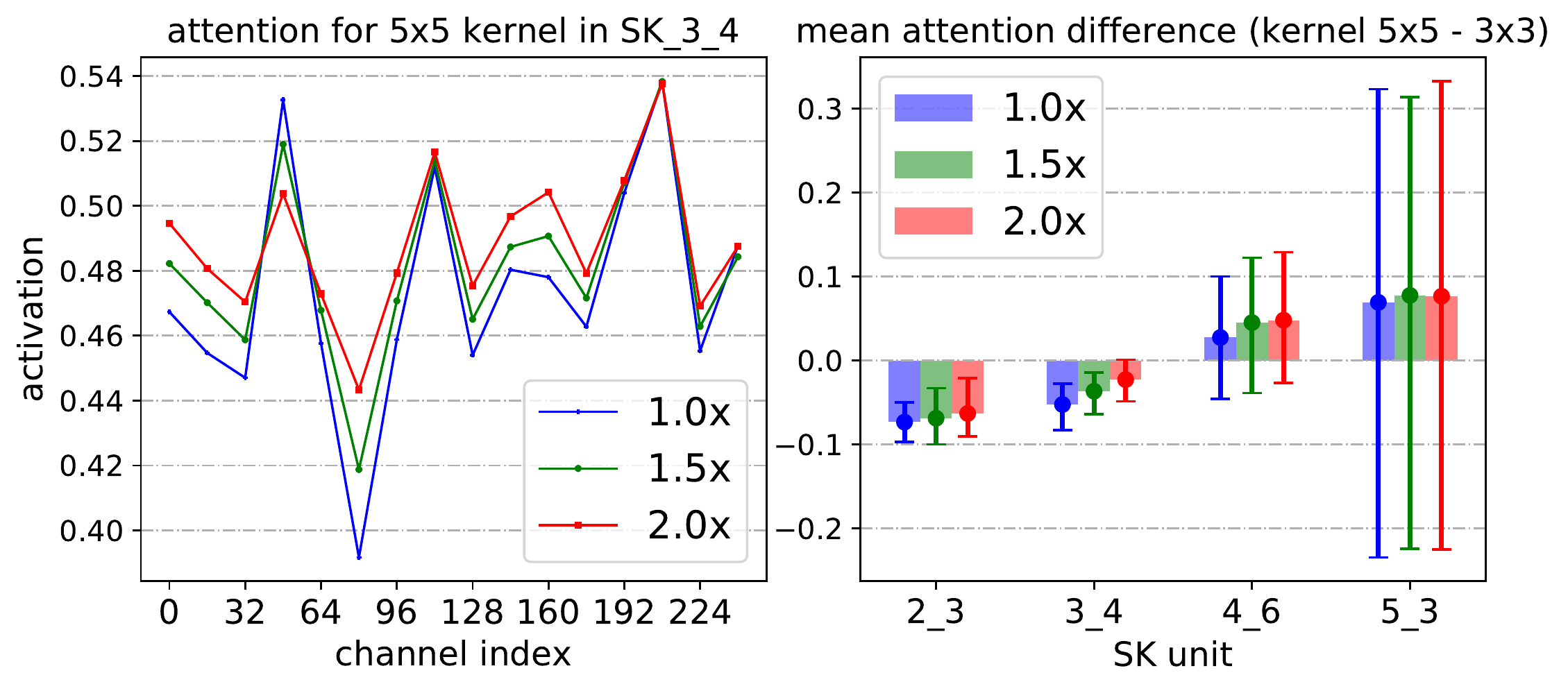}}
			\vspace{-4pt}
			
		\end{center}	
		\vspace{-4pt}
		\caption{(a) and (b): Attention results for two randomly sampled images with three differently sized targets (1.0x, 1.5x and 2.0x). Top left: sample images. Bottom left: the attention values for the 5$\times$5 kernel across channels in SK\_3\_4.  The plotted results are the averages of 16 successive channels for the ease of view. Right: the attention value of the kernel 5$\times$5 minus that of the kernel 3$\times$3 in different SK units. (c): Average results over all image instances in the ImageNet validation set. Standard deviation is also plotted.}
		\vspace{-10pt}
		\label{fig_show}
	\end{figure}

	\begin{figure*}[t]
		\begin{center}
			\setlength{\fboxrule}{0pt}
			\fbox{\includegraphics[width=\textwidth]{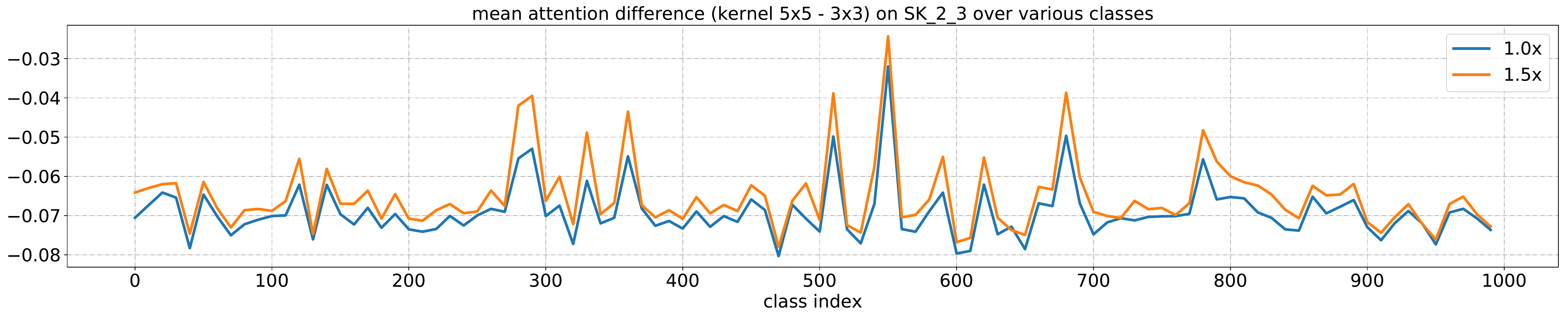}}
			\fbox{\includegraphics[width=\textwidth]{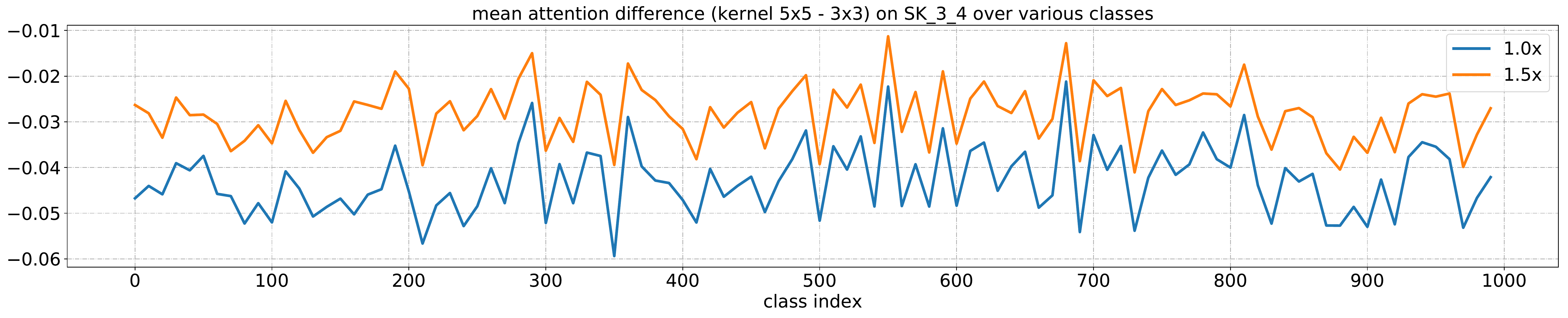}}
			\fbox{\includegraphics[width=\textwidth]{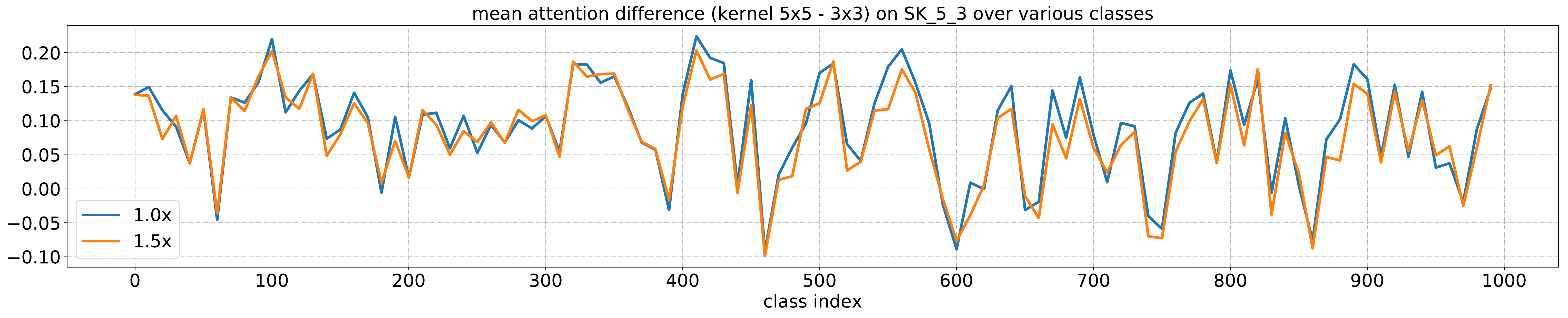}}
		\end{center}	
		\vspace{-12pt}
		\caption{Average mean attention difference (mean attention value of kernel 5$\times$5 minus that of kernel 3$\times$3) on SK units of SKNet-50, for each of 1,000 categories using all validation samples on ImageNet. On low or middle level SK units (e.g., SK\_2\_3, SK\_3\_4), 5$\times$5 kernels are clearly imposed with more emphasis if the target object becomes larger (1.0x $\to$ 1.5x). }
		\label{fig_all}
		\vspace{-4pt}
	\end{figure*}
	
	\subsection{Analysis and Interpretation}
	To understand how adaptive kernel selection works, we analyze the attention weights by inputting same target object but in different scales. We take all the image instances from the ImageNet validation set, and progressively enlarge the central object from 1.0$\times$ to 2.0$\times$ via a central cropping and subsequent resizing (see top left in Fig. \ref{fig_show}a,b). 
	
	First, we calculate the attention values for the large kernel (5$\times$5) in each channel in each SK unit. Fig. \ref{fig_show}a,b (bottom left) show the attention values in all channels for two randomly samples in SK\_3\_4, and Fig. \ref{fig_show}c (bottom left) shows the averaged attention values in all channels across all validation images. It is seen that in most channels, when the target object enlarges, the attention weight for the large kernel (5$\times$5)  increases, which suggests that the RF sizes of the neurons are adaptively getting larger, which agrees with our expectation.

	We then calculate the difference between the the mean attention weights associated with the two kernels (larger minus smaller) over all channels in each SK unit. Fig. \ref{fig_show}a,b (right) show the results for two random samples at different SK units, and  Fig. \ref{fig_show}c (right) show the results averaged over all validation images. We find one surprising pattern about the role of adaptive selection across depth: The larger the target object is, the more attention will be assigned to larger kernels by the Selective Kernel mechanism in low and middle level stages (e.g., SK\_2\_3, SK\_3\_4). However, at much higher layers (e.g., SK\_5\_3), all scale information is getting lost and such a pattern disappears.

	Further, we look deep into the selection distributions from the perspective of classes. For each category, we draw the average mean attention differences on the representative SK units for 1.0$\times$ and 1.5$\times$ objects over all the 50 images which belong to that category. We present the statistics of 1,000 classes in Fig. \ref{fig_all}. We observe the previous pattern holds true for all 1,000 categories, as illustrated in Fig. \ref{fig_all}, where the importance of kernel 5$\times$5 consistently and simultaneously increases when the scale of targets grows. This suggests that in the early parts of networks, the appropriate kernel sizes can be selected according to the semantic awareness of objects' sizes, thus it efficiently adjusts the RF sizes of these neurons. However, such pattern is not existed in the very high layers like SK\_5\_3, since for the high-level representation, ``scale'' is partially encoded in the feature vector, and the kernel size matters less compared to the situation in lower layers.

	\section{Conclusion}
	Inspired by the adaptive receptive field (RF) sizes of neurons in visual cortex, we propose Selective Kernel Networks (SKNets) with a novel Selective Kernel (SK) convolution, to improve the efficiency and effectiveness of object recognition by adaptive kernel selection in a soft-attention manner. SKNets demonstrate state-of-the-art performances on various benchmarks, and from large models to tiny models. In addition, we also discover several meaningful behaviors of kernel selection across channel, depth and category, and empirically validate the effective adaption of RF sizes for SKNets, which leads to a better understanding of its mechanism. We hope it may inspire the study of architectural design and search in the future. 
	
	\textbf{Acknowledgments}
	The authors would like to thank the editor and the anonymous reviewers for their critical and constructive comments and suggestions. This work was supported by the National Science Fund of China under Grant No. U1713208, Program for Changjiang Scholars and National Natural Science Foundation of China, Grant no. 61836014. 
	
	{\small
		\bibliographystyle{ieee}
		\bibliography{egpaper_final}
	}

\appendix
\renewcommand{\thetable}{S\arabic{table}}
\renewcommand{\thefigure}{S\arabic{figure}}
\section{Details of the Compared Models in Table 3}
We provide more details for the variants of ResNeXt-50 in Table 3 in the main body of the paper. Compared with this baseline, ``ResNeXt-50, wider'' has $\frac{1}{16}$ more channels in all  bottleneck blocks; ``ResNeXt-56, deeper'' has extra 2 blocks in the end of the fourth stage of ResNeXt-50; ``ResNeXt-50 (36$\times$4d)'' has a cardinality of 36 instead of 32. These three structures match the overall complexity of SKNet-50, which makes the comparisons fair.
	
\section{Details of the Models in Table 4}
For fair comparisons, we re-implement the ShuffleNetV2 \cite{ma2018shufflenet} with 0.5$\times$ and 1.0$\times$ settings (see \cite{ma2018shufflenet} for details). Our implementation changes the numbers of blocks in the three stages from \{4,8,4\} to \{4,6,6\}, therefore the performances and computational costs are slightly different from those reported in the original paper \cite{ma2018shufflenet} (see Table 4 in the main body of the paper for detailed results). In Table 4, ``+ SE'' means that the SE module \cite{hu2017squeeze} is integrated after each shuffle layer in ShuffleNetV2, ``+ SK'' means that each 3$\times$3 depthwise convolution is replaced by a SK unit with $M$ = 2 (K3 and K5 kernels are used in the two paths, respectively), $r$ = 4 and $G$ is the same as the number of channels in the corresponding stage due to the depthwise convolution.

Note that the 3$\times$3 depthwise convolution in the original ShuffleNet is not followed by a ReLU activation function. We verify that the best practice for integrating SK units into ShuffleNet is also without ReLU activation functions in both paths in each SK unit (Table \ref{tab_relu}). 

\begin{table}[h]
	\small
	\centering
	\renewcommand\arraystretch{1.1}
	\newcommand{\tabincell}[2]{\begin{tabular}{@{}#1@{}}#2\end{tabular}}
	\begin{tabular}{c|c|c}
		\hline
		\makecell[c]{K3\\+ ReLU ?}&\makecell[c]{K5\\+ ReLU ?}& Top-1 error (\%) \\
		\hline
		\cmark&\xmark&28.65\\
		\hline
		\xmark&\cmark&28.40\\
		\hline
		\xmark&\xmark&\bf{28.36}\\
		\hline
		\cmark&\cmark&28.49\\
		\hline
	\end{tabular}
	\vspace{+4pt}
	\caption{Influence of activation functions in two paths of SK units based on ShuffleNetV2 1.0$\times$. Single 224$\times$224 crop is used for evaluation on the ImageNet validation set.}
	\label{tab_relu}
	\vspace{-4pt}
\end{table}

\section{Details of the Compared Models in Figure 2}
We have plotted the results of some state-of-the-art models including ResNet, ResNeXt, DenseNet, DPN and  SENet  in  Figure 2 in the main body of the paper. Each dot represents a variant of certain model.  Table \ref{tab_imagenet} shows the settings of these variants, the numbers of parameters, and the evaluation results on the ImageNet validation set. Note that SENets are based on the corresponding ResNeXts. 

\begin{table}[h]
	\small
	\centering
	\renewcommand\arraystretch{1.4}
	\newcommand{\tabincell}[2]{\begin{tabular}{@{}#1@{}}#2\end{tabular}}
	\begin{tabular}{l|c|c}
		\hline
		Method & \#P & Top-1 error (\%)\\
		\hline
		ResNet-50 \cite{he2016deep} & 25.56M & 23.9 \\
		ResNet-101 \cite{he2016deep} & 44.55M & 22.6 \\
		ResNet-152 \cite{he2016deep} & 60.19M & 21.7 \\
		\hline
		DenseNet-169 (k=32) \cite{huang2017densely} & 14.15M & 23.8 \\
		DenseNet-201 (k=32) \cite{huang2017densely} & 20.01M & 22.6\\
		DenseNet-264 (k=32) \cite{huang2017densely} & 33.34M & 22.2 \\
		DenseNet-232 (k=48) \cite{huang2017densely} &55.80M&21.3\\
		\hline
		ResNeXt-50 (32$\times$4d) \cite{xie2017aggregated} & 25.00M & 22.2 \\
		ResNeXt-101 (32$\times$4d) \cite{xie2017aggregated} & 44.30M & 21.2 \\
		\hline
		DPN-68 (32$\times$4d) \cite{chen2017dual} & 12.61M & 23.7 \\
		DPN-92 (32$\times$3d) \cite{chen2017dual} & 37.67M & 20.7 \\
		DPN-98 (32$\times$4d) \cite{chen2017dual} & 61.57M & 20.2 \\
		\hline
		SENet-50 \cite{hu2017squeeze} &27.7M&21.12\\
		SENet-101 \cite{hu2017squeeze}&49.2M&20.58\\
		\hline
		SKNet-26 &16.8M&22.74\\
		SKNet-50 &27.5M&20.79\\
		SKNet-101 &48.9M&20.19\\
		\hline
	\end{tabular}
	\vspace{+4pt}
	\caption{{The top-1 error rates (\%) on the ImageNet validation set with single 224$\times$224 crop testing.}}
	\label{tab_imagenet}
	\vspace{-6pt}
\end{table}

\begin{figure*}[t]
	\begin{center}
		\setlength{\fboxrule}{0pt}
		\fbox{\includegraphics[width=0.48\textwidth]{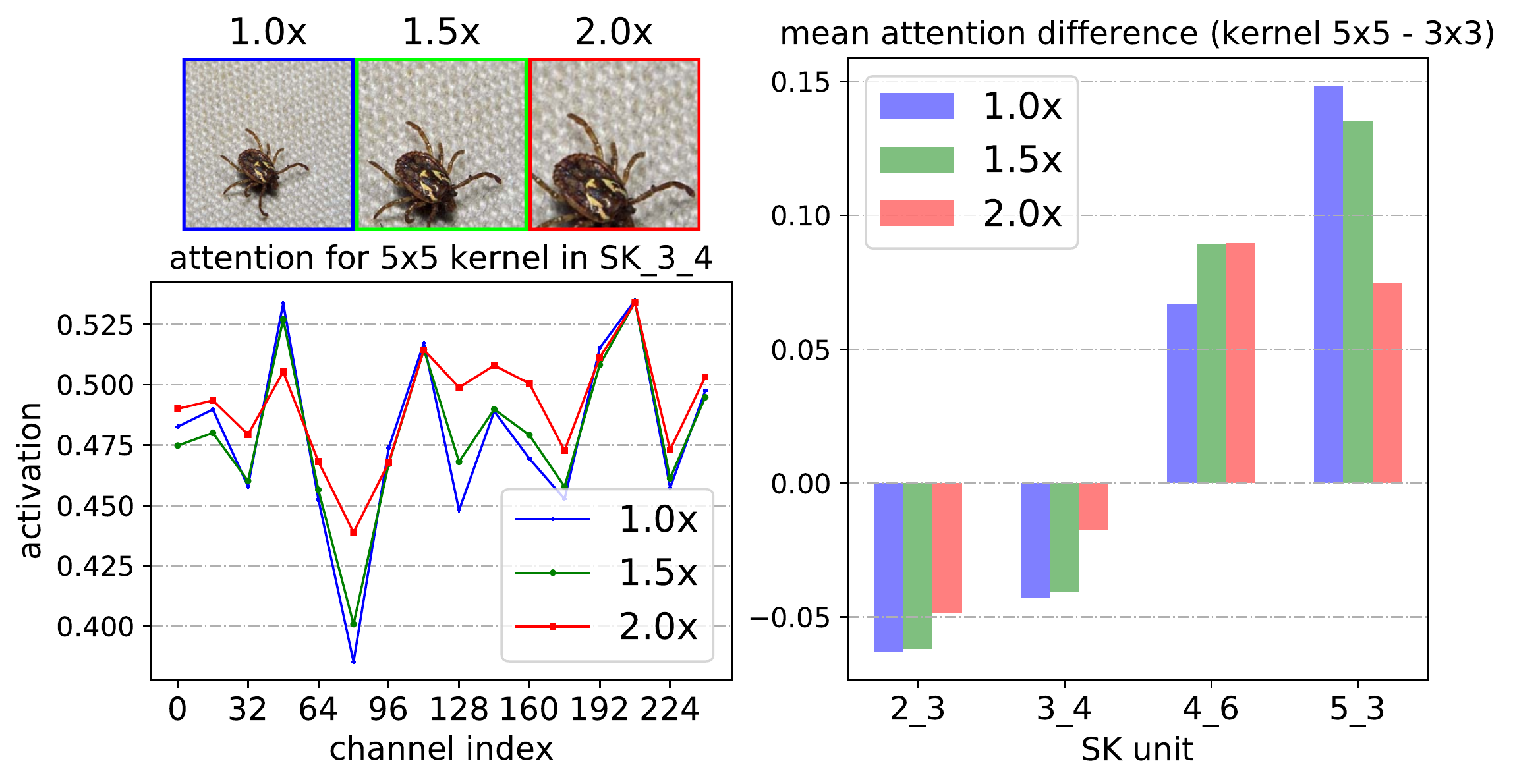}}
		\fbox{\includegraphics[width=0.48\textwidth]{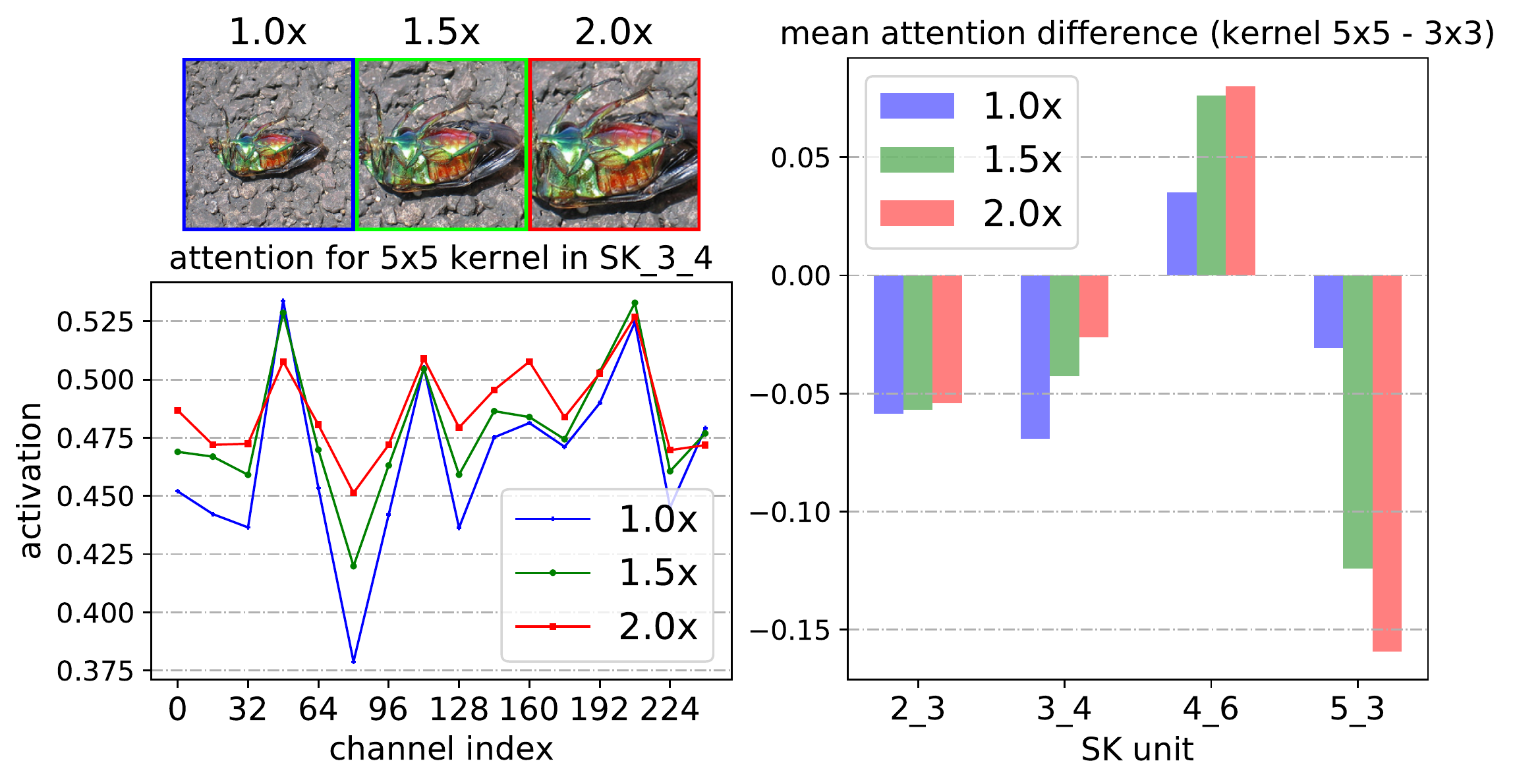}}
		\fbox{\includegraphics[width=0.48\textwidth]{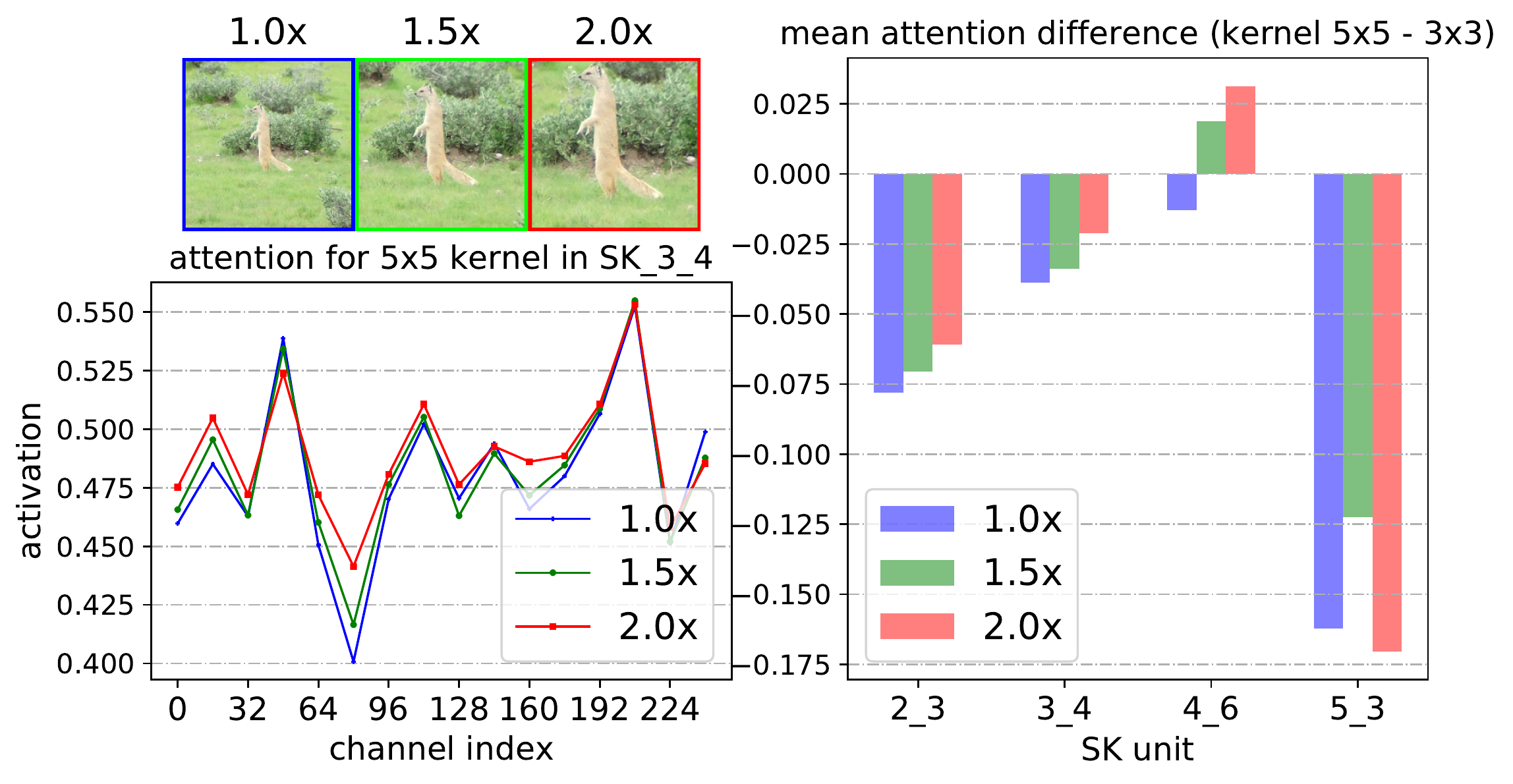}}
		\fbox{\includegraphics[width=0.48\textwidth]{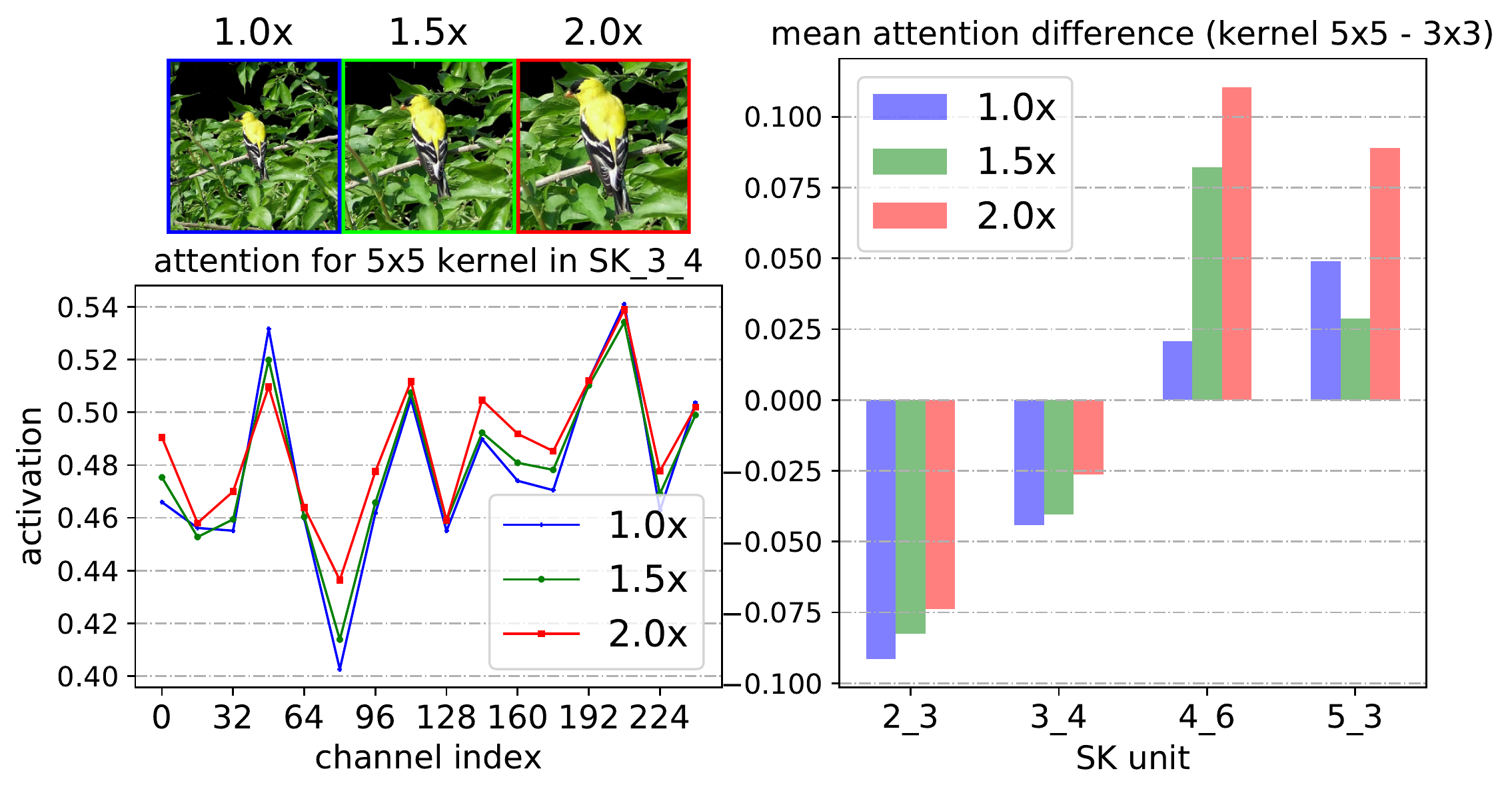}}
		
		\fbox{\includegraphics[width=0.48\textwidth]{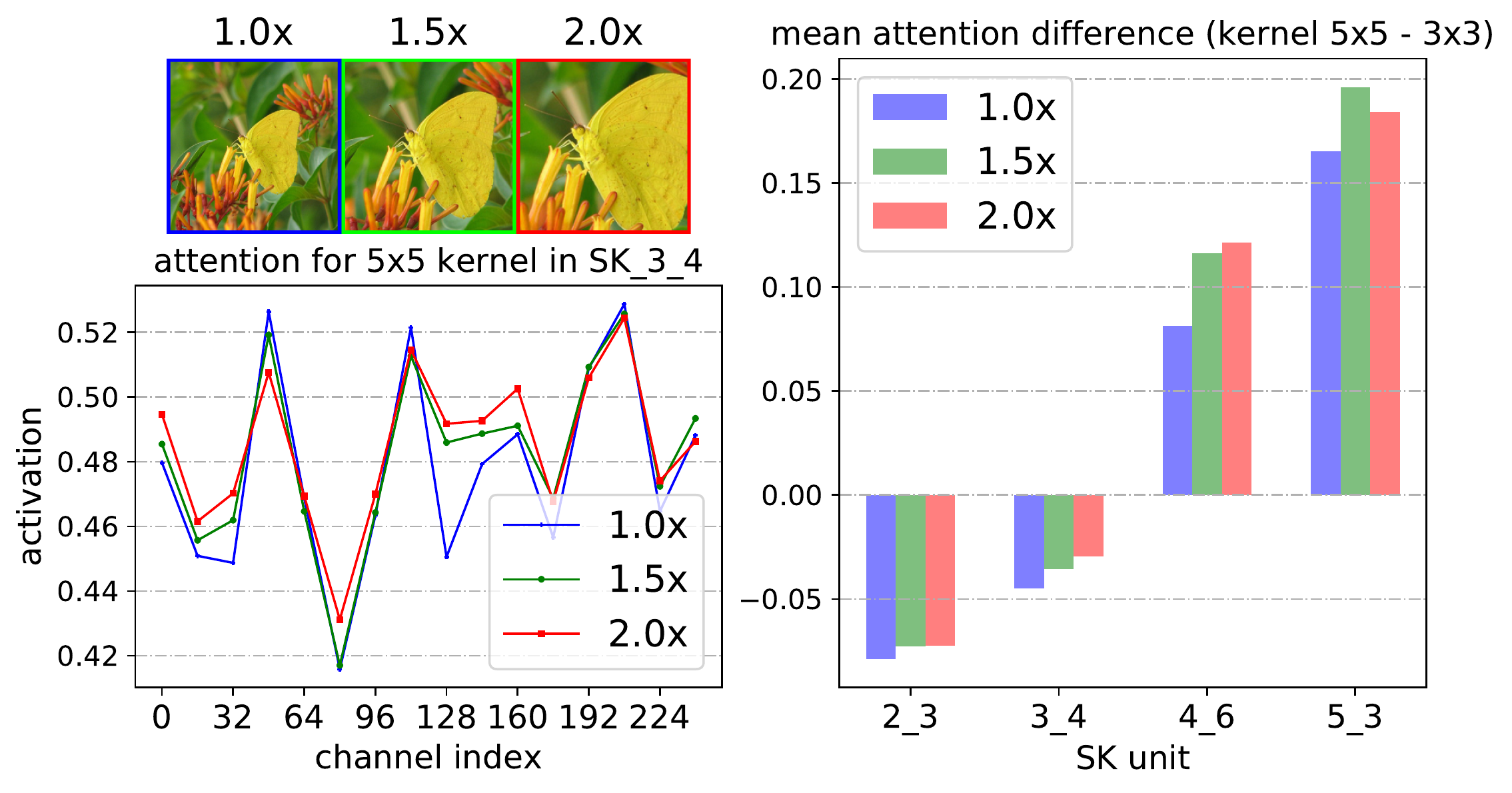}}
		\fbox{\includegraphics[width=0.48\textwidth]{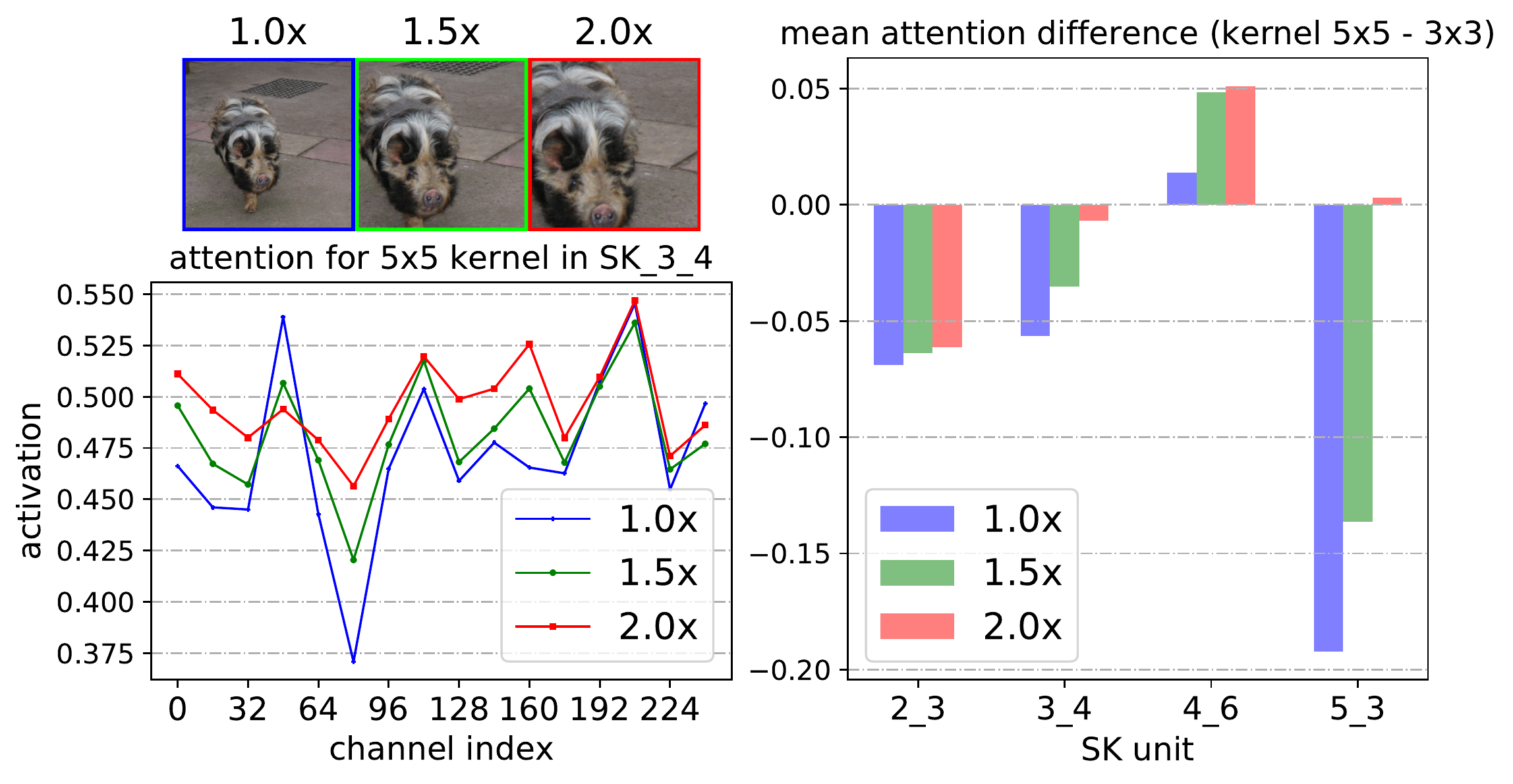}}
		
		\fbox{\includegraphics[width=0.48\textwidth]{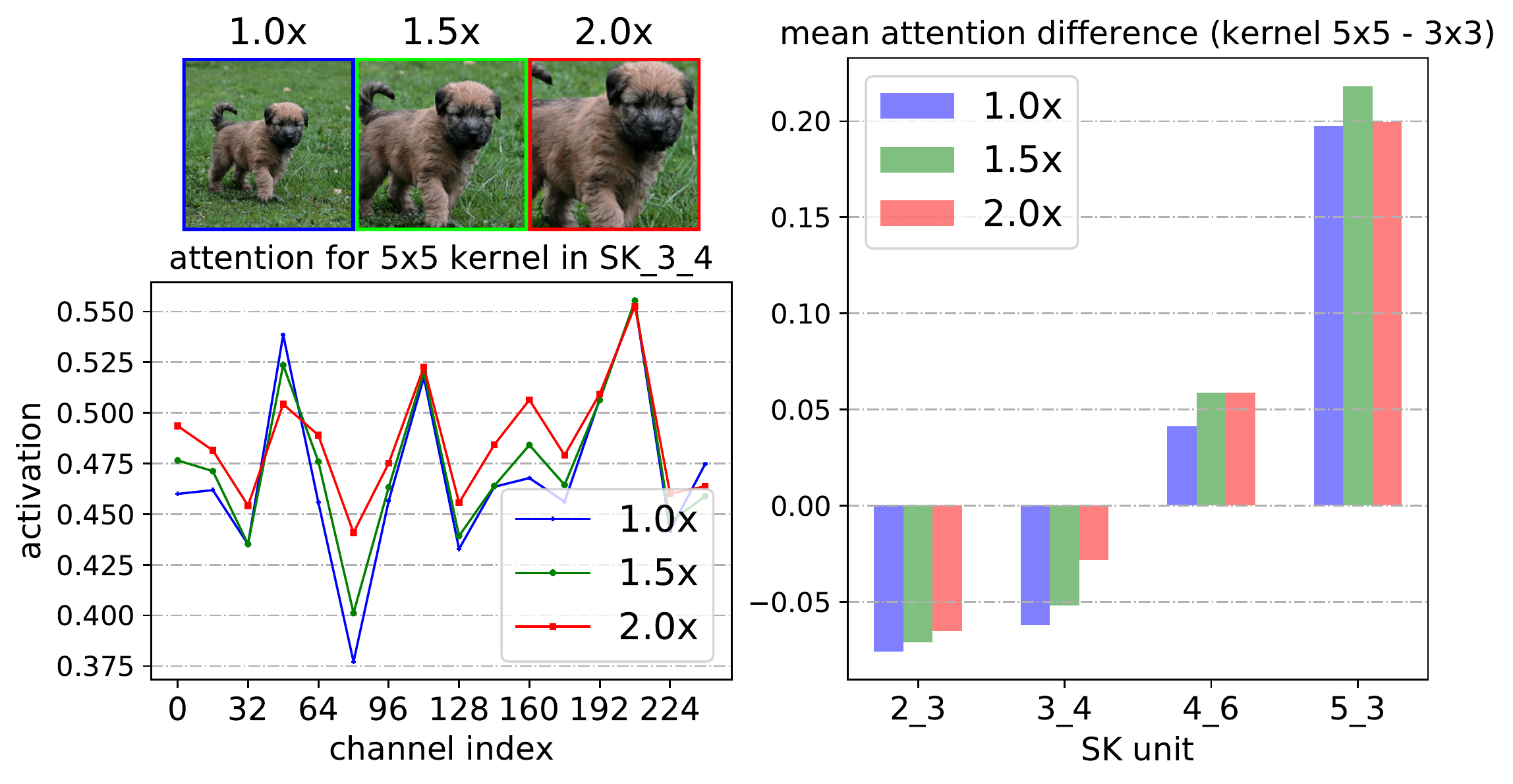}}
		\fbox{\includegraphics[width=0.48\textwidth]{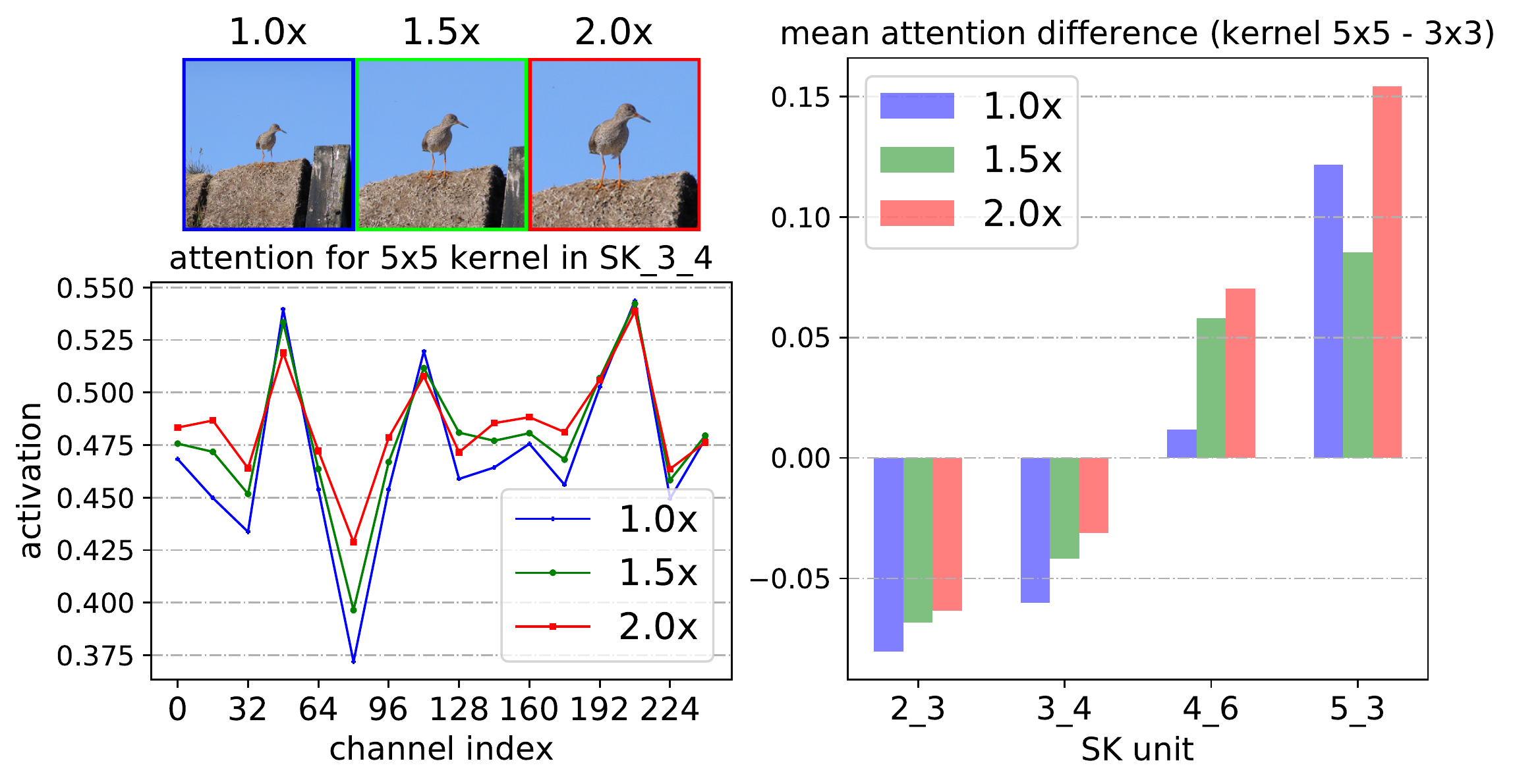}}
	\end{center}	
	\vspace{-4pt}
	\caption{Attention results for two randomly sampled
		images with three differently sized targets (1.0x, 1.5x and 2.0x). The notations are the same as in Figure 3a,b. }
	\label{fig_show_sk}
	\vspace{-4pt}
\end{figure*}

\section{Implementation Details on CIFAR Datasets (Section 4.2)}

On CIFAR-10 and CIFAR-100 datasets, all networks are trained on 2 GPUs with a mini-batch size 128 for 300 epochs. The initial learning rate is 0.1 for CIFAR-10 and 0.05 for CIFAR-100, and is divided by 10 at 50\% and 75\% of the total number of training epochs. Following \cite{he2016deep}, we use a weight decay of 5e-4 and a momentum of 0.9. We adopt the weight initialization method introduced in \cite{he2015delving}. The ResNeXt-29 backbone is described in \cite{xie2017aggregated}. Based on it, SENet-29 applies SE unit before each residual connection, and SKNet-29 modifies the grouped 3$\times$3 convolution to SK convolution with setting SK[2, 16, 32]. In order to prevent overfitting on these smaller datasets, we replace the 5$\times$5 kernel in the second path in the SK unit to  1$\times$1, while the setting for the first path remains the same.

\section{More Examples of Dynamic Selection}
Figure \ref{fig_show_sk} shows attention results for more images with three differently sized targets. Same as in Figure 3 in the main body of the paper, we see a trend in low and middle level stages: the larger the target object is, the more attention is assigned to larger kernels by the dynamic selection mechanism.

\end{document}